\documentclass[smallextended]{svjour3} 
\smartqed  
\usepackage{graphicx}
\usepackage{subfig}
\usepackage{multirow}
\usepackage{booktabs}
\usepackage{natbib}
\usepackage{tabularx}
\usepackage{url}

\newcommand{\tf}{TFIDF}
\newcommand{\kl}{$\textit{KLD}_\varepsilon$}
\newcommand{\cs}{$1-\textit{Cosine similarity}$}
\newcommand{\dvr}{DVR}
\newcommand{\pvr}{PVR}
\newcommand{\ps}{LPA-signature}
\newcommand{\pss}{LPA-signatures}
\newcommand{\lpa}{LPA}
\newcommand{\dln}{$L1-Norm$}
\long\def\omitit#1{}

\begin{document}

\title{Domain-based Latent Personal Analysis and its use for impersonation detection in social media}
\titlerunning{LPA: Latent Personal Analysis}    

\author{Osnat Mokryn \and Hagit Ben-Shoshan}
\institute{O. Mokryn \at Information Systems \\ University of Haifa\\Israel\\\email{ossimo@gmail.com}
\and H. Ben-Shoshan \at Management \\ University of Haifa \\ Israel\\ \email{hagits@gmail.com}
}
\date{Received: date / Accepted: date}
\maketitle
\begin{abstract}
 Zipf's law defines an inverse proportion between a word's ranking in a given corpus and its frequency in it, roughly dividing the vocabulary into frequent words and infrequent ones. Here, we stipulate that {\em within a domain} an author's signature can be derived from, in loose terms, the author's missing popular words and frequently used infrequent-words. We devise a method, termed Latent Personal Analysis (\lpa), for finding domain-based attributes for entities in a domain: their distance from the domain and their signature, which determines how they most differ from a domain. We identify the most suitable distance metric for the method among several and construct the distances and personal signatures for authors, the domain's entities. The signature consists of both over-used terms (compared to the average) and {\em missing} popular terms. We validate the correctness and power of the signatures in identifying users and set existence conditions. We test \lpa~in several domains, both textual and non-textual. We then demonstrate the use of the method in explainable authorship attribution: we define algorithms that utilize \lpa~ to identify two types of impersonation in social media: (1) authors with sockpuppets (multiple) accounts; (2) {\em front users} accounts, operated by several authors. We validate the algorithms and employ them over a large-scale dataset obtained from a social media site with over 4000 users. We corroborate these results using temporal rate analysis. \lpa~ can further be used to devise personal attributes in a wide range of scientific domains in which the constituents have a long-tail distribution of elements. 
\keywords{Latent Personal Analysis (LPA) \and Zipf \and Authorship attribution \and Impersonation \and Sockpuppets   \and Front users}
\end{abstract}

\section{Introduction}
The near-Zipfian nature of word frequencies is a well known and highly researched universal law. A communicative optimization explanation for its origin builds upon the principle of least effort~\citep{zipf1949human,i2003least}. The principle is rooted in a trade-off arising from the use of words. Frequent words (as reflected by their count in large corpora) are easier to choose, produce, and use~\citep{brown1966tip,akmajian2017linguistics}. On the other hand, the more frequent a word is the more meanings it has, and the more ambiguous it is, resulting in a need to enrich the communication with uncommon, contextual words~\citep{ferrer2018origins,Hahn2347}. 

Semantics strongly influences word popularity, as meaning is a substantial determinant of frequency~\citep{piantadosi2014zipf,calude2011we}.  Across domains, popular words may vary in meaning, and hence in frequency. This variance in word frequency was utilized as an underlying factor in topic modeling and latent semantic analysis~\citep{blei2003latent,dumais2004latent}. Within a domain, then, words' distribution can be described as having a {\em head}, consisting of popular domain words, and a {\em long tail} of a supporting vocabulary of infrequently used words. 
Personal language usage, however, varies between and within categories. As such, it has been frequently used for authorship attribution and verification. People vary in the richness of their vocabulary, choice of words, and style~\citep{brinegar1963mark,argamon2009automatically,schwartz2013personality}. Often overlooked, though, is the question of which words they do not choose.

The missing, not conveyed, is an integral part of the whole.  "Music is the silence between the notes", described Debussy; ~\cite{freud1925negation} coined negation as the dual process to affirmation,  an expulsion from one's perception of reality. The part that is present, yet not recognized:  "It is now no longer a question of whether something perceived (a thing) shall be taken into the ego or not, but of whether something which is present in the ego as an image can also be re-discovered in perception (that is, in reality)". In language, we claim, style is the silence between the words. Domain-popular words that are not used should contribute to personal characterization. We maintain that considering the words users choose to use and those they do not use would help identify their style. A choice to omit a domain-popular term is just as characteristic of personal style as a choice to include a domain-infrequent one. 

Building on the above observation, we hypothesize that personal signatures in a domain can be derived by defining how a person's vocabulary {\em differs} most from the domain's vocabulary. But how does one determine this difference? If we negatively assume that there are no personal differences, then a personal vocabulary could be considered a random sample from the domain vocabulary. This would yield a minimal information loss when measured by a relative entropy distance, e.g., Kullback-Leibler Divergence~\citep{Kullback1951}.
 By contradiction, then, personal differences are the elements that contribute most to the distance measured between a personal vocabulary and the domain.

To find the distance metric best suited for the task, we compare the resulting distance distributions from four chosen distance metrics over a very large corpus. The examined metrics are drawn from different worlds: The first two are used to measure the similarities between texts indefinite ranked lists. Both are then adapted to measure distance; The other two are the  Euclidean distance and an epsilon-padded Kullback-Leibler (\kl~) distance. We apply three selection criteria that distinguish quantitatively between the four distance metrics, yielding \kl~ as the chosen distance metric.

Using \kl, we construct two attributes for each entity: (1) its distance from the domain,  which is a measure of its uniqueness in the domain. (2) its domain-signature, a vector representation with the word-elements contributing most to the entity's distance from the domain. The signature is a vector representation of what most {\em separates} an entity from the domain.

We then demonstrate the uses of both attributes for exploring entities' characteristics in a textual domain. Here, we term each entity 'a writing', and the elements are words. 
 
When the domain consists of a single author's writings, we show that each writing (entity) distance should be relatively small. Hence, distant writings could be suspected as unoriginal. 
Similarly, when the domain consists of various authors, each author's typical distance (entity) is relatively large. We then show that the text of entities with a small distance to the domain is written by several authors. We use this finding to uncover this type of impersonation, which we term {\em front users}, a user account operated by several authors, in a social media site.    

The signature is a short vector representation of the entity,  summarizing how it is different from the domain.  We demonstrate that it contains contextual information and validate that similarity between signatures indicates similarity in style. We show how this can be used in different settings for authorship attribution. 

The method is easy to compute as it is distance-based and does not require learning while offering explainability.  Hence, we explore its applicability for online social media sites.  

In recent years, social media and User Generated Content (UGC) have revolutionized our lives and habits~\citep{kaplan2010users}. Users are no longer passive consumers of content but active and contributing participants. They create, share, and engage with online content ~\citep{kaplan2010users,van2009users}.  Strangers' opinions are considered, at times, more important and influential than those obtained from close friends and appreciated sources~\citep{kietzmann2011social}.  The coupling of this tendency with online platforms' anonymity has led to the rise of false and fake information, promoted by bots, fake accounts, and impersonation~\citep{ferrara2016rise,vosoughi2018spread}.

We identify two types of impersonations in social media.
One form of impersonation is the prevalent case of multiple accounts belonging to a single author, allowing users to have multiple online identities ("sockpuppets"). The second one is what we term {\em front-users}. Front-users are online accounts used by a group of users that utilize the account as a front user persona. Front-users accounts were found in the case of political and influence groups, for example. 
We obtain over a million reviews of roughly 4000 users from IMDb, a movie review platform, and identify several hundreds of sockpuppets and front-user accounts. The list of sockpuppets accounts was sampled and validated qualitatively; Accounts suspected of being front-users were analyzed for their temporal activity patterns for validation. Analytical and visual results are given that corroborate our findings, and we compare the performance of \lpa~exploration in social media for identifying authors of multiple accounts with a vector-based baseline. 

We further demonstrate how \lpa~can be used in a non-textual context by exploring countries' song listenings using a dataset obtained from Spotify. 
\lpa~can be used in various fields in which the constituting entities have a long-tail distribution of elements, such that a clear separation of a head that contains domain-popular elements a tail of less frequent elements is constructed as the domain-frequency ranked list. The domain-based signatures can be used to model entities for content-analysis, recommender systems, and other modeling tasks.

\section{Data used in this research}
\label{sec:data}
To validate and test our method, we need large corpora of written texts. These corpora, our domains, must be large enough for statistical tests and must be composed of many people's writings. We use two such datasets: 1) the Books dataset, consisting of 30 English language books, and 2) the Social media dataset, composed of the collected reviews of 3,969 IMDb users. The Books dataset is a set of labeled data that we will use to validate our method. It contains 30 English language books, taken from the Gutenberg project's most popular books list. The IMDb dataset is an example of the kind of data our method was designed to handle, and we will use it to demonstrate its applicability. It contains 1,406,000 movie reviews, spanning the period of July 1998 - June 2016. Each review contains a text, a timestamp, and an author ID. The original obtained IMDb dataset contained $467,961$ users. To have a large enough text sample for each user, we extracted only users who published at least 30 reviews. $3,969$ users met this criterion. Appendix A details the characteristics of the datasets.

\begin{figure}[!ht]
\centering
      \subfloat[Reviews per user\label{imdbstats-1}]{%
       \includegraphics[width=1.1\textwidth]{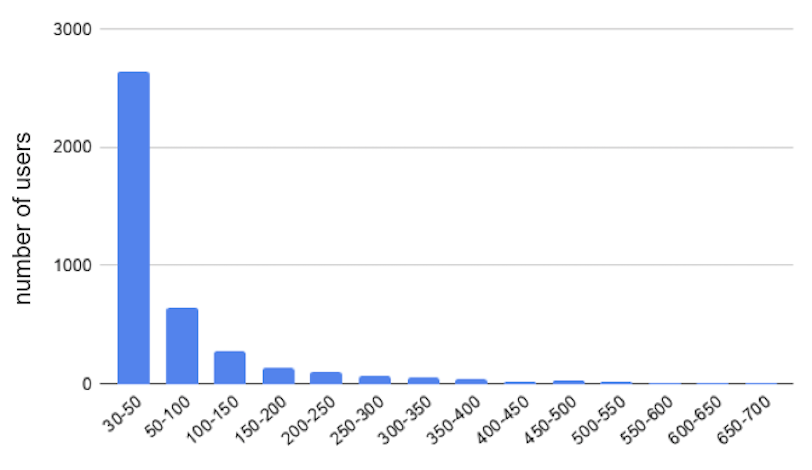}
     }\\
     \subfloat[Users' vocabulary size \label{imdbstats-2}]{%
       \includegraphics[width=\textwidth]{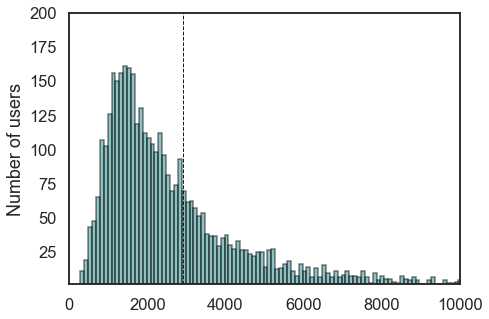}
     }
     \caption{IMDb dataset statistics: (a) Distribution of the number of reviews per user. (b) Distribution of users' vocabulary sizes. The vertical dashed line denotes the average vocabulary size}
     \label{fig:imdbstats}
   \end{figure}

Table~\ref{tbl:imdbrevs}  lists relevant statistics for the IMDb dataset. Figure~\ref{fig:imdbstats}(a) details the distribution of written reviews per author (user). Figure~\ref{fig:imdbstats}(b) shows the distribution of vocabulary sizes, with the dashed vertical line denoting the average vocabulary size. The contribution of the number of reviews per user is right-skewed, with the majority of users writing less than 200 reviews. 
\subsection{Missing popular terms}
\begin{figure}[!ht]
\centering
     \subfloat[Percentage of missing terms from top 1000\label{imdb-mis-1}]{

       \includegraphics[width=0.95\textwidth]{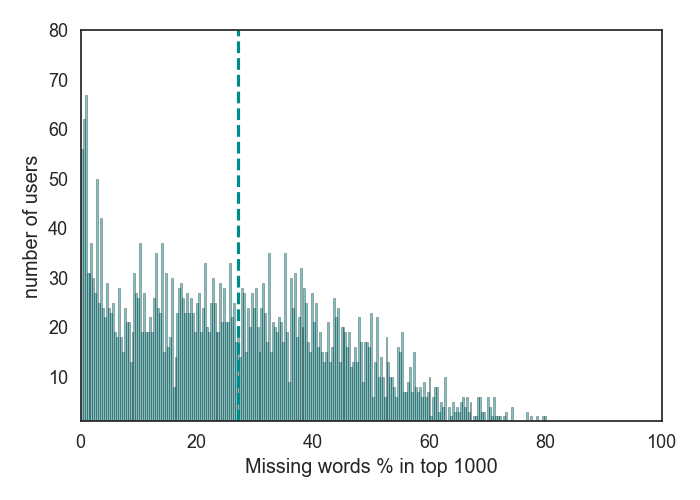}
     }
\\
     \subfloat[Percentage of missing terms normalized by vocabulary size\label{imdb-mis-2}]{
       \includegraphics[width=0.95\textwidth]{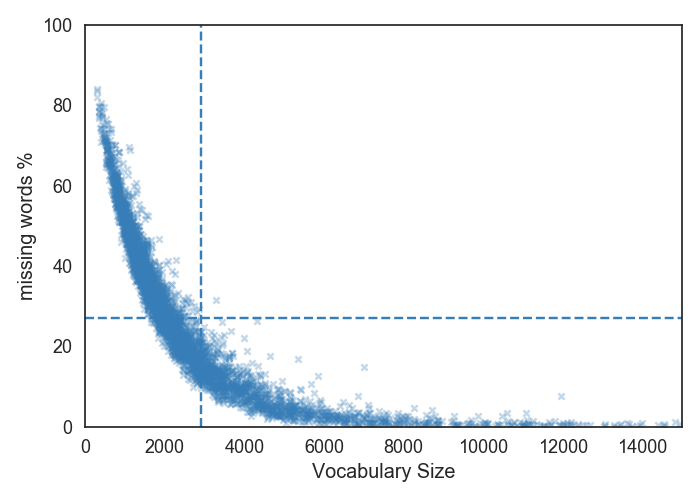}
     }
     \caption{IMDb dataset: missing terms in users' vocabulary. (a) Distribution of missing top 1000 domain terms from users vocabularies. (b) Distribution of missing top 1000 domain terms normalized by the users' vocabulary size.}
     \label{fig:imdb-mis}
   \end{figure}

Our method would also consider popular terms that are missing from a user's vocabulary. To motivate the use of missing terms, we detail here relevant statistics per user over the IMDb film domain. Figure~\ref{fig:imdb-mis}(a) shows the distribution of missing domain popular terms from the vocabulary of review authors. The figure was calculated for users authoring more than 30 reviews, and it depicts the distribution of the number of popular terms, out of the most popular 1000 terms in the domain, that is missing from their vocabulary. Figure~\ref{fig:imdb-mis}(b) shows the percentage of missing terms, normalized by the vocabulary size. The vast majority of users do not use terms that are among the top 1000 in the domain. The number of missing popular terms decreases as the vocabulary of the user is bigger. However, there is no linear relationship between them, and most users with 2000 and more terms in their vocabulary miss only 25\% of the popular terms or less. This indicates that popular terms are prevalent in the use of most users. Hence, the lack of such use of a prevalent term should be indicative and distinct. 

\section{Domain-based Latent Personal Analysis method}
For a set of entities that constitute a domain, LPA intends to find the entities' signatures, defined as the elements that make the entities unique and set them apart from the domain.  In language, that would be the words that set authors apart from the domain. In music, signatures can be the songs that set a user apart; if the entities considered are countries, their signatures would consist of the songs that set countries' listening habits apart.
As the method's insight arrives from languages' characteristics, we describe it here using text and authors. To find how a person's vocabulary differs most from the domain's vocabulary, we need to establish which distance metric is best suited for the task.

Here, we will (1) explain the setup; (2) Devise criteria for choosing a distant metric for the method and explore which metric, out of four chosen candidates, is most suitable for our needs; (3) Demonstrate that the resulted entities' distances from the domain's vector are not an artifact of them being part of the domain but are significant, and also encode important information that can be utilized; (4) Devise a criterion for choosing signature lengths and construct \pss; (5) Give examples of the construction process for the Books and Social Media datasets, and also an example for non-text domains taken from the music industry. 

\subsection{\lpa~ setup}
Here, we use the following definitions. Considering the used datasets, described in Section~\ref{sec:data}: Each book, chapter, or user's reviews are entities in a domain.  When referring to the Books dataset, an entity is either a book or a chapter, according to need. For the social media dataset IMDb, hereto referred to as Social Media dataset, an entity is a user, and the corresponding entity's text is the aggregation of all the reviews written using a single user account. The collection of all the texts by all authors in each dataset is termed a domain document. Each entity's text is termed an entity's document. 
We take a Bag of Words (BoW) approach to deal with the unstructured text: tokenization, case folding and normalization, punctuation and stop-words removal, part of speech tagging. 
(We omit stop-words since we want the signatures to characterize the entities for explanatory and content-based characterization).
We create a weighted vector for the domain and each entity's document. 
The weighted domain vector is termed Domain Vector Rates (\dvr). Each weighted entity's vector is termed Personal Vector Rates (\pvr). For the construction demonstration and definition we use the noun-terms. 

\subsubsection{\dvr~ construction}
The weighted collection vector, \dvr, is constructed out of the entire collection, the aggregation of all the domain entities. In the case of a book, the collection is the entire aggregated text of all the chapters. In the Social Media dataset, it is the entire text written in all reviews by all users considered in the domain. 

To construct the \dvr, we compute each word's relative weight as its relative frequency out of all the domain words. For example, a word that took up 10\% of all users' BoW is assigned a weight of $0.1$. 
In the following toy example, we take a small part of the Social Media dataset to demonstrate the method setup. Table~\ref{tbl:dvr-example} depicts the elements, their global frequency in the collection, and their corresponding \dvr~ weights.  The most popular word is ranked first in this list, etc. 
\begin{table}[h!]

  \caption{Toy example of a domain vector construction}
  \label{tbl:dvr-example}
  \centering
  \begin{tabular}{ l c c}
    \toprule
   Noun  &    Global frequency &    Global probability \\
     term &     in collection&    in collection\\
    \midrule
Film&    1276332&    0.0212\\
Movi&    1117427&    0.0184\\
Charact&    406469&    0.0067\\
Time&    379878&    0.0062\\
Stori&    339747&    0.0056\\
  \bottomrule
\end{tabular}

\end{table}
\subsubsection{\pvr~ construction}
Similarly, we construct a weighted frequency vector per entity. It should be noted that each entity uses a smaller vocabulary than the entire set of entities in the corpus. For example, if a book is a domain, and each chapter is an entity, the number of different words at each of the chapters is probably smaller than the number of distinct words in the book. Similarly, each user's number of words in all of their reviews in the Social Media dataset is probably much smaller than the number of different words used by all of them together in what we term the domain. That is, most authors do not use every word in the \dvr. However, we would like to compare their vectors with that of the domain. Therefore, we construct entities' vectors as sparse vectors, containing null coordinates for words existing in the corpus but not in the entity's writings. An entity's \pvr, therefore, is usually very sparse. An example of a few such random rows for our Toy example is shown in Table~\ref{tbl:author-example}.
\begin{table}[h!]

  \caption{Toy example of an entity vector construction}
  \label{tbl:author-example}
  \centering
  \begin{tabular}{ l l}
    \toprule
Word&    Probability\\
\midrule
favorite&    0.546\\
drama&    0.318\\
film&    0.136\\
boat&    Null (0)\\
brother&  Null (0)\\
  \bottomrule
\end{tabular}
\end{table}
\subsection{Distance metric selection}
The choice of a distance metric is central to the method. If chosen correctly, then the elements most contributing to the distance of an entity from its domain are the ones that distinguish this entity. Per our discussion, these elements are elements that the entity contains much less relative to the domain and much more. Here, we use the Social Media dataset. 
\subsubsection{Distance metrics considered}
\label{sec:4dist}
Distance metrics attempt to formalize the natural properties of spatial distance. They are generally used to introduce a geometric structure to more abstract sets. For this research, however, we are not interested in such a geometric structure.  Specifically, as we are always interested in the distance between two distinct points, and not in the shortest way to reach one from the other, we are willing to consider functions which do not satisfy the Triangle inequality. Such functions are called semi-metrics. However, here, we will informally use the term {\em metric}.
 We define here the distance metrics we chose. In Appendix B, we prove that each of our selected metrics is indeed a distance metric.
Reviewing past research in the subject and related areas, we selected four candidates to serve as our metrics. In Appendix B, we show that each is a distance metric, i.e., fulfills the following criteria: non-negativity, symmetry, and identity of indiscernibles (proofs omitted from here for lack of space). 
\paragraph{Rank Biased Overlap Distance (RBD)}
Rank Biased Overlap, or RBO~\citep{Webber2010a}, assigns a similarity value to non-conjoint lists, i.e., lists which do not necessarily have the same items. RBO assigns a similarity value to two such lists by calculating their overlap over various segments of the lists. It is rank-biased because overlapping at the head of the lists contributes to the similarity value more than overlap at the tail. RBO itself is not a metric but a similarity measure. However, we can derive a distance metric from it, termed RBD, by defining $RBD=1-RBO$.

To calculate the RBO value of two lists $V_1,V_2$, we first consider the intersection size between the lists over the first $d$ terms, i.e., how many elements appear in the first $d$ terms of both lists. We call this value $X_d$. We then define $A_d$, the agreement between the two lists over the first $d$ terms, as $A_d=\frac{X_d}{d}$. Note that $X_d\leq d$ and therefore $A_d\leq1$.
Then the RBO similarity measure of the two lists $V_1,V_2$ is defined as follows:
\begin{equation}
    RBO(V_1,V_2,p)=(1-p)\sum_{d=1}^\infty p^{d-1}\cdot A_d
    \label{eq:rbo}
\end{equation}
Where $p$ is a parameter in the range $0<p<1$ which determines the weight given to the head of the lists - the closer $p$ is to zero, the greater the bias towards the head. 
The distance between the two lists is then $RBD=1-RBO$. 
\paragraph{Cosine Similarity}
Cosine similarity between two vectors $V_1, V_2$ is defined as follows: 
\begin{equation}
    D(V_1,V_2)=\frac{V_1\cdot V_2}{\| V_1 \| \times \| V_2 \|}
    \label{eq:cosine}
\end{equation}
Where $\cdot$ is the standard dot product, $\times$ is the standard multiplication and $\|v\|$ is the standard $L^2$ Euclidean norm. 

\paragraph{\dln}
The \dln~ has been in use since at least 1757 when Roger Joseph Boscovitch used it for regression analysis. Since then, it has often been used in assessing the difference between discrete frequency distributions, making it a good candidate for our needs. 
For $V_1(x)$, $V_2(x)$ frequency vectors, we used a modified version of  \dln~ distance such that the maximal distance between the lists is 1. Let $V_1$ and $V_2$ be non-conjoint lists, then
\begin{equation}
    L_1(V_1,V_2)=\frac{1}{2}\sum_{x\in X} \arrowvert V_1(X)-V_2(X)\arrowvert
\end{equation}

\paragraph{Kullback-Leibler Divergence (KLD) with a back-off}
The Kullback-Leibler Divergence (KLD)~\citep{Kullback1951} is a familiar statistical method for measuring the difference between two distributions $V_1$ and $V_2$.
For two distributions, $V_1,V2$, of a set of elements $X$, KLD is calculated by:
\begin{equation}
D([V_1\Arrowvert V_2])=\sum_{x\in X}\Bigg[\big[V_1(x)-V_2(x)\big]log\Big[\frac{V_1(x)}{V_2(x)}\Big]\Bigg]
\label{eq:kl}
\end{equation}
KLD was designed for two distributions of the same set of elements, as it assumes that for every $X$, both $V_1(x),V_2(x)$ are non-zero. In our case, however, most users (entities) do not use the domain's entire vocabulary. That is, for some elements $x \in X$ we have $V_1(x)=0$. 

To account for the large variance in vector lengths and counterbalance the domain's long-tail of rare words, we use a variant, the KLD with a back-off model, defined in~\cite{bigi2003using}.  This variant assigns missing words a constant value $\epsilon$. This value is chosen to fulfill two requirements: it must be less than the frequency of a single word in the entire domain (corpus), and it must be large enough so that KLD still gives results in the range $[0,1]$. In effect, we expand the user's (entity's) vector from a sparse vector to one containing a positive value in every coordinate. This, however, also means that the extended vector, $V_e$, is no longer a probability vector - the sum of all coordinates is now larger than 1. We correct this by multiplying all non-$\epsilon$ frequencies by a normalization coefficient $\beta$. This normalization coefficient is given by the formula $\beta=1-N\epsilon$, where N is the number of words missing in one vector compared to the other. 

Thus, an extended frequency vector $V_e$ is defined by: 
\begin{equation}
V_e(t_k\| d_j)=
\left\{\begin{array}{lr}
\beta V_1(t_k\| d_j)  & t_k \in d_j  \\
     \epsilon  &\quad \textit{otherwise} \\ 
\end{array}\right\} 
\end{equation}
Where $d_j$ is the original vector before the extension, not containing missing words. 

We can verify that this is indeed a frequency vector: let $X$ be the set of all indices in the vector, $K\subset X$ the set of all non-$\epsilon$ indices, i.e. those that had non-zero value in the original vector and $N$ the number of missing words, i.e. how many coordinates have an $\epsilon$ value. We then have \[\sum_{x\in X}V_e(x)=\sum_{x\in K}V_e(x)+\sum_{x\not\in K}V_e(x)=\sum_{x\in K}\beta V_1(t_k\| d_j)+\sum_{x\not\in K}\epsilon\] but since $V_1$ is a frequency vector, $\sum_{x\in K} V_1(t_k\|d_j)=1$ and we therefore have \\
\[\sum_{x\in X}V_e(x)=\beta \sum_{x\in K} V_1(t_k\|d_j)+N\epsilon=1-(N\epsilon)\times 1 + N\epsilon=1\] 
We refer hence-forth to the~\citep{bigi2003using} version as \kl.

\subsubsection{Selection criteria}
We compare here the performance of the distance metrics according to four criteria. A distance metric used in our method should: 
\begin{enumerate}
    \item Provide wide distribution of results when calculating entities' distance from the domain, indicating that the metric is sensitive enough to small differences between entities. We used Kurtosis as an indicator of the distribution's width. Kurtosis is a statistical measure that defines how the tails of a given distribution differ from the tails of a normal distribution. 
    A wide distribution should have relatively small Kurtosis Excess. 
    \item Missing popular words should contribute to the distance. 
    \item The distance function should be more sensitive to differences at the head of the vector (the relatively few, more popular words) than at its tail. Vocabularies tend to have a long-tail distribution, and this criterion ensures that many minor differences in the tail would not overshadow the few, more significant, differences at the head.
\end{enumerate}
\subsubsection{Criterion 1: small kurtosis excess of distances from the \dvr}
We calculate each entity's distance from the \dvr~ in each of the four proposed metrics. 
\begin{table*}[t!]
  \caption{Criterion 1 results: distance distribution statistics including Kurtosis Excess for each of the distance metrics used}
  \label{tbl:crit1}
  \begin{tabularx}{\textwidth}{c@{\extracolsep{\fill}} c c c c c c}
    \toprule
    Metric&Min&Max&Average&Std.&Median&kurtosis Exc.\\
    \midrule
    RBD &4.49E-08&02.05E-06&8.481E-08&2.819E-08&7.95E-08&6.51\\
    1- cos.-sim. &0.00034&0.04561&0.003039&0.00246&0.00249&2.8\\
    $L_1$ &0.237383&0.775756&0.435991&0.069562&0.433481&-1.2\\
    \kl &0.035 &0.861&0.223&0.114&0.203&1.48\\
  \bottomrule
\end{tabularx}
\end{table*}
Table~\ref{tbl:crit1} depicts the distance distribution statistics for each of the distance metrics. The distribution is the distribution of users' vectors' distances from the \dvr. Of the four, RBD (recall that $RBD=1-RBO$) has the highest kurtosis excess and the narrowest distance distribution. Therefore, we reject RBD and continue with the other three: \cs, \dln~and \kl.

\omitit{
\begin{figure}[h]
 \centering
 \includegraphics[width=.95\textwidth]{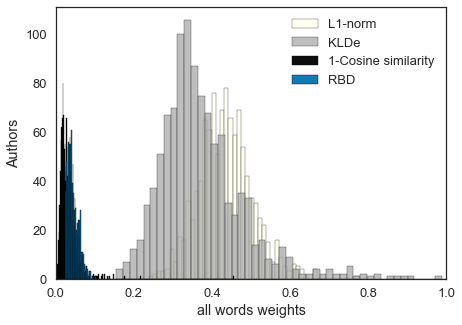}
 \caption{MISSIng CAPTION DO NOT USE  }
 \label{fig:criter1}
\end{figure}
}
\subsubsection{Criterion 2: contribution of missing popular items to the distance} 
To measure the contribution of missing popular words, we select the most popular 1,000 terms in the domain. We then calculate each entity's distance from the domain, based on these 1,000 terms, and determine the contribution of each user's missing words to that distance, as calculated by the three remaining distance metrics. 
\cs~ fails to measure this contribution, as only terms that appear in both vectors with values greater than zero add to the distance. Hence, we reject the use of \cs, and are left with \dln~ and \kl.  
\subsubsection{Criterion 3: distance function sensitive to differences in the head and less sensitive to differences in the tail}
\label{sec:criter3}
\begin{figure}[!ht]
\centering
\includegraphics[width=\textwidth]{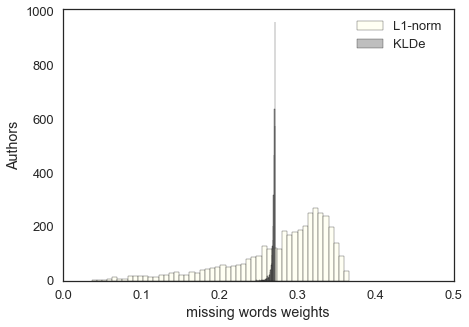}
     
  \caption{Criterion 3: the contribution of missing words in the tail to overall distance under \kl~ and \dln.}
  \label{fig:criter3}
   \end{figure}

To test the compatibility of the remaining metrics to this criterion, we took the complement of the vectors from the previous one - here, we only considered the terms after the $1,000$ most popular. Figure~\ref{fig:criter3} depicts the distributions for this criterion.
\omitit{
\begin{figure}
\subcaptionbox{Criterion 2: visualization of the contribution of frequent missing words to the distance distribution of users from the \dvr~ measured for \dln, \cs, and \kl}{ \includegraphics[width=.95\textwidth]{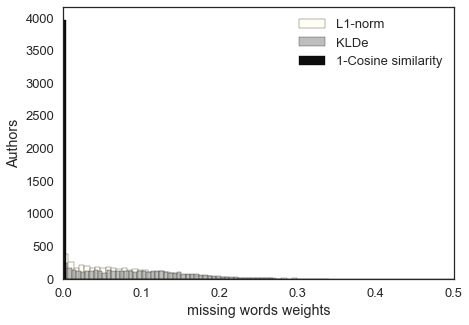}}%
\\
\subcaptionbox{Criterion 3: the contribution of missing words in the tail to overall distance under \kl~ and \dln. }{\includegraphics[width=.95\textwidth]{criterion3_BW.png}}%
\caption{Visualization of criterion 2 on the left panel and of criterion 3 on the right panel, yielding the choice in \kl~ as the method distance metric}
\label{fig:criter23}
\end{figure}
}

The results show that \dln~ is biased to the right (larger distance). This is consistent with the fact that it correlates with the entity's vocabulary size - an author with a large vocabulary will have fewer missing words and, therefore, be closer to the \dvr.
As \dln~ assigns a high value to rare missing terms, it fails this criterion, and therefore \kl, which provides a wide distribution and a balanced average, is our chosen metric. 
We then wrap-up the method construction by utilizing \kl~ as the chosen metric.

\subsection{Validity of the \kl~distance as an attribute of an entity}
We continue to validate that an entity's distance from the \dvr~is not merely an artifact of the entity being a sample from the domain. We choose 210 users from the Social Media dataset; each contributed 100 to 120 film reviews of different lengths. We term them 'authentic' users.
We then create 210 virtual users, whose text was chosen randomly from the dataset. We select randomly for each virtual user between 100 to 120 reviews at random from the dataset.
We then create signatures for both authentic and virtual users and compute the distances. 
Let $f_n$ be the distribution of distances of normal users from the \dvr. Let $f_v$ be the distribution of distances of virtual users from the \dvr. Then, we hypothesize that there is a significant difference between $f_n$ and $f_v$.
We use an independent-sample t-test to compare the two distributions and find a significant (t(418)=-59.05, p=0.00) difference between the authentic users (M=0.459, SD=0.11) and the virtual users (M=0.919, SD=0.14). 
This allows us to reject the null hypothesis and confirm that authentic authors' distance distribution is significantly different from that of a randomly selected text of similar size. 

\textbf{In summary}, \lpa~distances are a newly found attribute of the entities.  An entity's \lpa~distance is a measure of how similar is its distribution to that of the domain's. Our criteria require that entities with a similar head of distribution to that of the domains' are closer to the domain (short distance) than entities that differ substantially in the elements constituting the head of their distribution. When the entities are users and the elements are terms, users who use the popular domain terms have a shorter distance to the domain than those who do not.  We will discuss the outcomes of this observation in Section~\ref{sec:multiple}. 

\subsubsection{Constructing LPA signatures}
We have established a distance measure based on how an entity is different from the domain. However, as we will show here, for many purposes, entities can be represented by their signature, a vector representation constructed of the top N terms most contributing to the entity's \lpa~distance from the domain. That is, what makes entities unique, and sets them apart from the domain. Utilizing the entity's \pvr~ distance from the \dvr, we continue to form an entity's signature as the N-words which {\em contribute most} to this distance. 
Two points should be noted regarding this definition: i) A signature is not a trait of a particular text, but a text relative to another text. A signature is a vector representation constructed out of the N-words, which most differentiate a given text from a given domain. So while we will use the term \textit{signature} here, it should always be considered relative to a specific domain. ii) A signature might contain elements that do not exist for the entity; in our case, words that are not in the author's vocabulary. If a specific word is prevalent in the domain but absent from an author's writings, that tells as much about the author as if a word was very infrequent in the collection but very popular in the author's writings.

\paragraph{Working set size}
Works using word frequency vectors share a common difficulty in determining the working set's size.~\cite{bigi2003using}, in her KLD back-off model work,  used a working set of 500 terms for text categorization.~\cite{baker1998distributional}, likewise working on text classification, have achieved 66\% accuracy with the top 500 words per text. Researching corpus similarity measures, ~\cite{kilgarriff1998measures} have achieved the best results when using the 320-640 most popular terms. ~\cite{ntoulas2006detecting} used the most frequent 500 words to detect spam webpages. They Found that 500 words were enough to extract 75\% of the relevant information.
 By the very definition of a signature, the N-words which contribute most to the user's distance from the \dvr, each additional word should have a diminishing return on the distance. Therefore, we want to use as few words as possible while still retaining the essential characteristics reflected by the distance. We proceed to show that $N=500$ indeed justifies these claims.

\begin{figure}[!ht]
 \centering
 \includegraphics[width=1.1\textwidth]{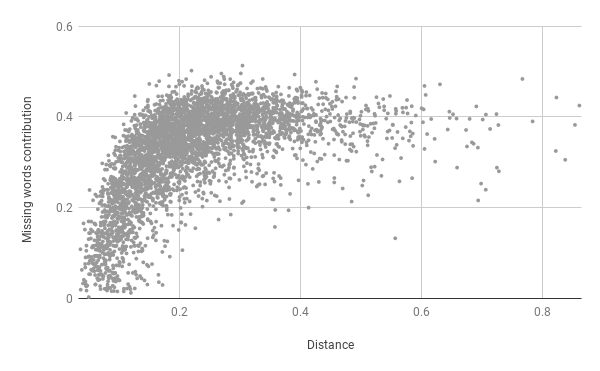}
 \caption{Correlation between entity's distance from the \dvr~and the contribution of missing terms to the distance}
 \label{fig:latent-trend}
\end{figure}

We analyzed the two datasets,  Books and Social Media. Table~\ref{tb:sigsize} details the average calculated contribution of the top furthest 50, 100, 500, and 1,000 words to the distance in two datasets. In both datasets, the top 500 words contribute around 50\% of the distance. A further increase to the top 1,000 words yields an increase of only 10\% of the total distance. When considering the diminishing return of additional words, we find that the diminishing return was less than one percent around the $100_{th}$ word. The exact length of the signature is then a trade-off between computing effort and precision.  Based on the above-described findings, we also set the signature size to 500, i.e., $N=500$. 
\begin{table}[t!]
\centering
  \caption{The average contribution of the top 50, 100, 500 and 1,000 words  to a user's vector distance from the \dvr}
  \label{tbl:signature-size}
  \begin{tabular}{ c c c c c}
    \toprule
Dataset&    50 & 100& 500& 1,000\\
\midrule
Social Media&    16.32\% & 22.70\% & 45.84\% & 55.11\% \\
Books & 19.87\% & 29.28\% & 56.80\% & 66.20\% \\
  \bottomrule
\end{tabular}
\label{tb:sigsize}
\end{table}
We further see that the first $N=500$ terms contribute to the distance differently to different entities. For some, the top N words contribute 70\% and more of the measured distance. To others, the top N words contribute less than 50\% of their \kl~ distance. Specifically, we find that the top N words contribute less to entities closer to the \dvr~ and more to distant entities. In other words, the more distinct a user's style is, the more their signature is representative of that style.  This result implies that taking the top N terms gives a robust result across all users.
\paragraph{Latent terms contribution to the signatures}
Using our Books dataset, we create signatures for 1,118 chapters relative to the entire dataset. We measure the distance of these signatures from the \dvr~ and examine what percentage of each signature's distance is contributed by words existing in the user's vocabulary and what percentage is contributed by words missing from it. Figure~\ref{fig:latent-trend} shows the correlation.
As expected, the results show a clear trend. That is, for distant entities, a larger percentage of the distance was contributed by missing words. 

\subsection{An entity's LPA signature: the essence of an entity's difference from the domain}
\label{sec:examples}
We demonstrate LPA-signatures~ construction with Daniel Defoe's novel Robinson Crusoe (1719) and the Social Media dataset. We also give a brief example of how \lpa~ can be used in various domains using a dataset of song popularity per country released by Spotify in 2017. 

 The story of Robinson Crusoe is of an Englishman, castaway on a deserted island for 28 years, later saved to return to civilization. He returns accompanied by Friday, an escaped prisoner he rescued\footnote{The information regarding the book is taken from ~\cite{ben2008robinson} and https://en.m.wikipedia.org/wiki/Robinson\_Crusoe.}. 
\begin{figure}[!ht]
\vspace{-.1in}
\centering
     \subfloat[Top terms in the signatures of Chapters 1, 10, and 20, corresponding to the first chapter, half-way, and the last one. 
     \label{fig:robinson1}]{

       \includegraphics[width=0.95\textwidth]{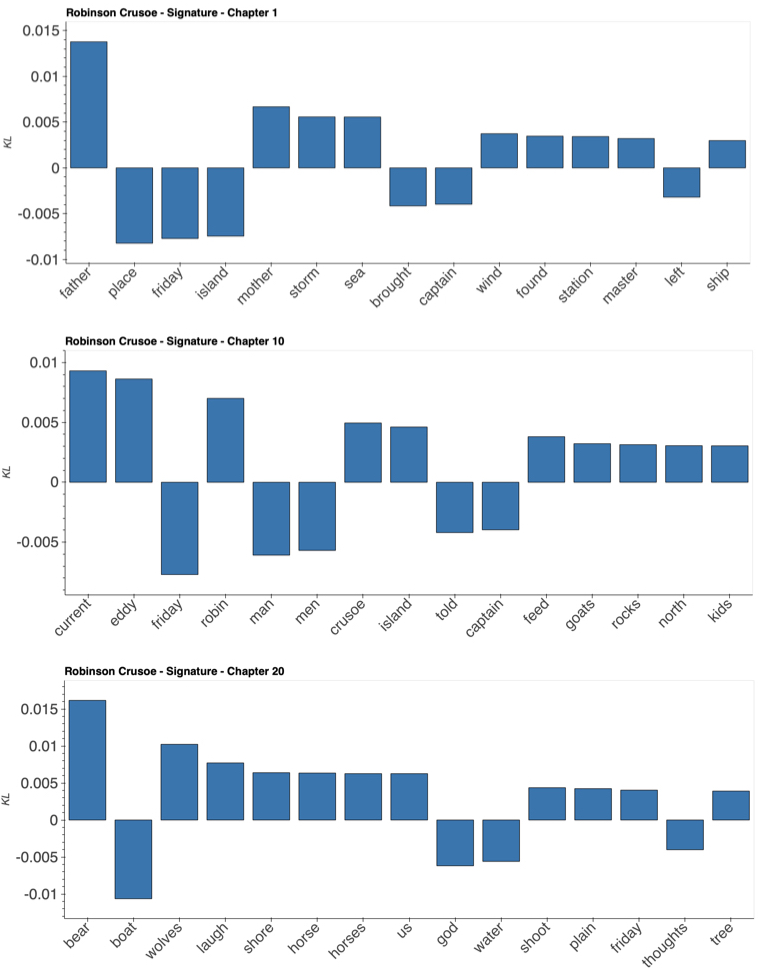}
     }
\\
     \vspace{-.1in}
     \subfloat[The lower panel depicts the novel's top \dvr~terms. On the upper panel, the top terms from Chapter 16's signature. In this chapter, Friday and Robin rescue Friday's father and plan to leave the island.\label{fig:robinson2}]{
       \includegraphics[width=0.95\textwidth]{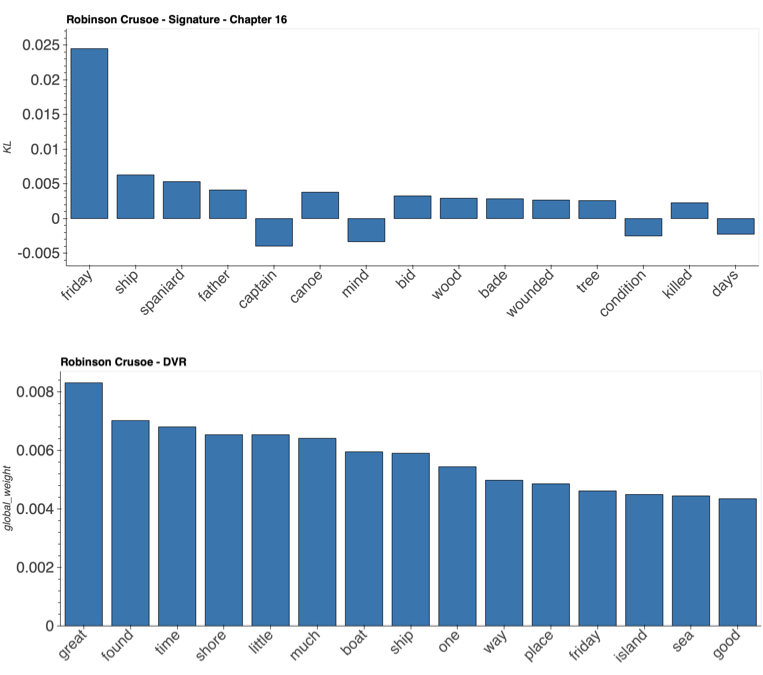}
     }
     \caption{ Daniel Defoe's novel Robinson Crusoe: \dvr, \pss}
     \label{fig:curose}
   \end{figure}

In the construction of this novel with \lpa, the entities are the chapters, and the domain is the book. Thus, the \dvr~shows the novel's term frequency, while the \ps~of each chapter depicts the terms that most set the chapter apart from the novel.  Figures~\ref{fig:robinson1}, \ref{fig:robinson2}, show the domain's top \dvr~terms and the top terms of the \pss~ of chapters 1 (the first), 10, 16, and 20 (the last).  The novel's top twenty terms consist of prominent plot-related terms such as shore, boat, ship, island, sea, Robinson's companion Friday, and God.  In chapter one, the parents play a central role, as well as the sea. In chapter ten,  other people (man, men) are discussed much less. Instead, the parrot echoing Robin's name is prominent, as is Robin's mishap in the stream. Chapter 16 is dominated by Friday, the Spaniard, and the term "mind" is underused. The final chapter, 20, concentrates on a fight on the shore with a bear. The prominent missing terms are boat, God, water, thoughts. The term laugh, however, appears much more than in the rest of the book. 

\begin{figure}[!th]
 \centering
 \includegraphics[width=\textwidth]{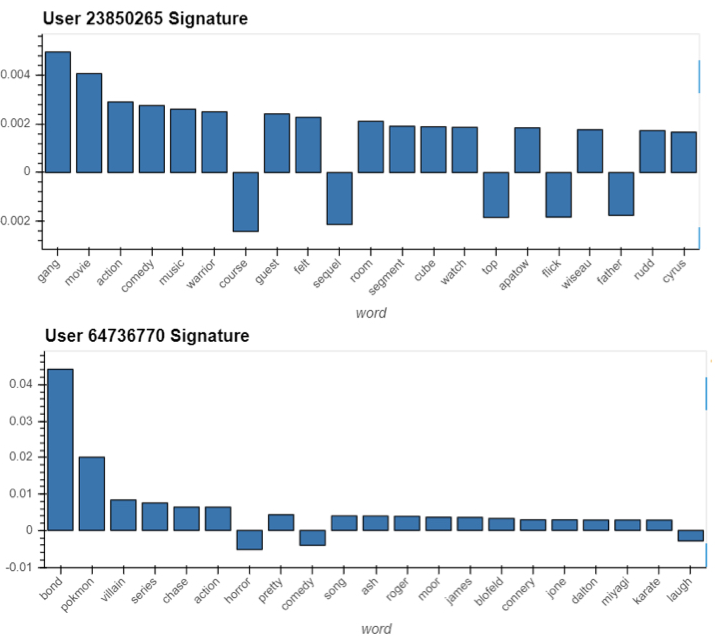}
 \caption{\pss~ of two users from the Social Media dataset, a film review platform. Signatures contain both overused and underused terms.}
 \label{fig:imdb-sampl}
\end{figure}
For the Social Media dataset, we give the \pss~of two randomly selected users. Figure~\ref{fig:imdb-sampl} Shows their signatures. We see on the upper panel a user discussing movies, thoughts, and how the user 'felt'. They talk less about cast, script, and horror films than is common. On the lower panel, we see a Bond films' reviewer who avoids horror and comedy and mentions the word 'laugh' less than is common. 

\begin{figure}[!th]
 \centering
 \includegraphics[width=\textwidth]{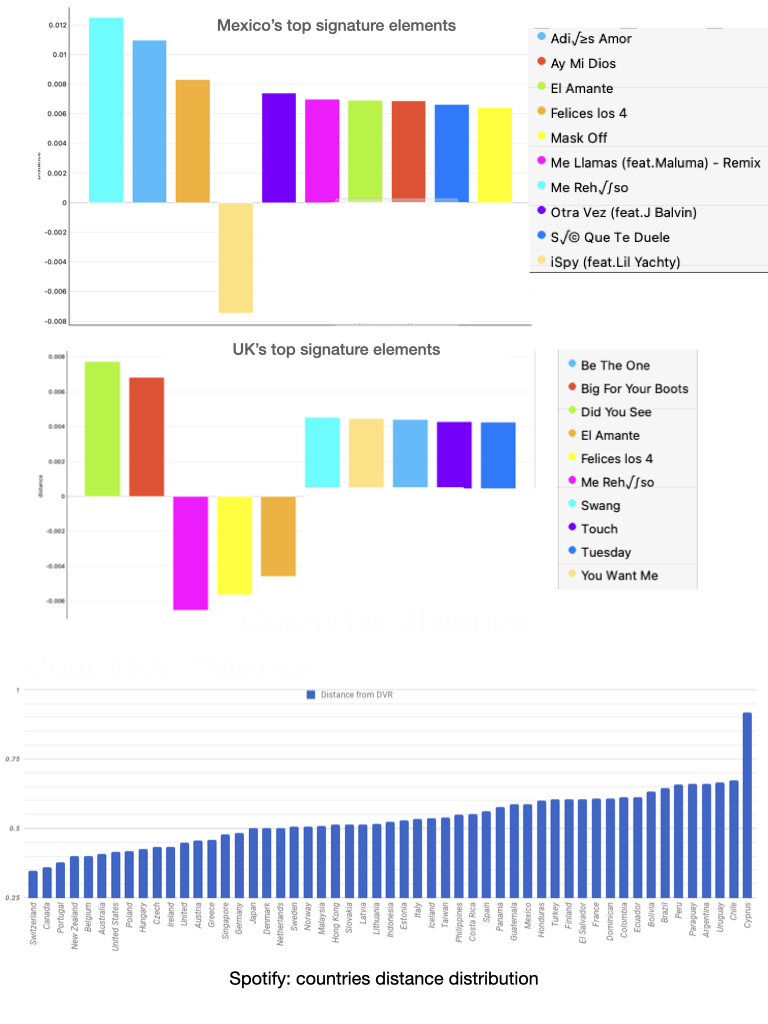}
 \caption{LPA-signatures for the music domain: an example over a Spotify dataset from 2018. The lower panel gives the distances of countries from the \dvr. The upper panel details the top elements in the \pss~for Mexico (left upper panel) and the  UK (right upper panel). The UK is closer to the \dvr, and the head of its distribution is similar to the top world view; hence its signature contains subtler changes. In Mexico, local music is more favored; hence Mexico is more distant, and its signature weights are larger than the UK's.}
 \label{fig:spotify}
\end{figure}
 
 To demonstrate how \lpa~can be used in non-textual domains, we take as an example a dataset of songs from Spotify from 2017. The dataset contains all songs streamed in that year and their listening frequency in each country. To create the \dvr, we created a vector frequency of the aggregated countries' lists, obtaining a world frequency view of songs' popularity. We then create the \pss for the countries. Figure~\ref{fig:spotify} shows The results. The lower panel depicts the distance from the \dvr~ and relative stream size for each country.  Countries that listen more to popular music are closer to the \dvr. Examples of these are Canada, Switzerland, and Australia. These countries' signatures are characterized by small differences in listening habits to hit songs. Countries distant from the \dvr~are these in which local music is preferred, for example, Turkey or Uruguay. On the upper panel, we see the signatures of two countries with a roughly equal volume of listened tracks, Mexico and the UK. Mexico, the distant one, has a signature that is dominated by popular local music tracks. The UK is closer in distance to the world and has a signature that is dominated by subtle differences from the \dvr. 

\subsection{\lpa~calculation: a fast and slim process}
Let N be the total number of terms, or elements, in the domain, and Q the total number of entities in a domain. 

\textbf{Memory considerations}: To construct \pss, the algorithm requires that entities' vector lengths be identical to that of the \dvr's.  The total memory size for this phase is $3\cdot N \cdot Q$, considering each element has a corresponding weight, and after calculating the \kl~distance, a sign. However, the working set size consists of 500 terms (signature size) per entity, and thus the memory requirement for the \pss~ is merely $500 \cdot Q$. 

\textbf{Execution time considerations}: There are two phases of computations for the \pss~construction. In the setup phase, the \kl~distance between each entity and the \dvr~is computed. The distance between each entity's vector and the \dvr~is given by Equation~\ref{eq:kl}, and requires one pass over the vectors' length while computing a log function. Hence, computing the distances is of O(N) complexity. In the second phase, the \pss~are computed by choosing the top 500 terms. 

\textbf{Similarity time considerations}: Computing the similarity between any two entities' signatures requires an $L_1$ computation over 500 elements. That is, it is $O(Q^2)$. 
When used as a similarity measure, that is, finding similar signatures, \lpa~requires computation power of the order of the number of entities considered to the power of 2 (that is $Q^2$).   

\section{LPA's uses}

\lpa~ creates two representations for each entity in a specific domain: the entity's distance from the domain and its signature (vector representation) in the domain.  

Here, we show that an entity's distance from the domain is a measure of its uniqueness. In the case of text, it is a measure of style. When the domain consists of a single author's writings, each sample's distance (that is, an  \lpa~entity) from the domain is small. However, when the domain consists of various authors' writings, a relatively small distance of a sample text (and \lpa~entity) is an indication that multiple authors are responsible for that text, perhaps pretending to be a single author. 

The signatures are a vector representation of an entity in a domain. Hence, in the case of text and authorship, when two entities' vectors' representations are very similar, it is an indication that a single author wrote both text samples (entities). We further show that the representations of two chapters of the same book are closer to each other than to chapters' representations of other books by the same author. Further, when the domain consists of various authors' writings, the similarity between books' vector representations (\pss) is a measure of styles' closeness. 

We conclude this section by considering the amount of information needed to produce reliable signatures by comparing sampled chapters' signatures to complete chapters' signatures.  

Here, we will mostly use the Books dataset since this is a fully labeled dataset: each chapter is part of a book written by a single author. As will be discussed, we consider each book both as a domain and an entity, dependent on the context.  Thus, conditional on the experiment, either the book is an entity, or a chapter is an entity. When the chapters are entities, the experiments can contain hundreds of entities. 

\subsection{\lpa~ entities' distances as a measure of authorship}
Here we show that the \lpa's entities distances can identify authorship. We show that book chapter entities are closer to their book's \dvr~than to arbitrary books' \dvr's. We create arbitrary books, which we term 'virtual' books, by randomly choosing chapters from various books written by different authors. 
We construct each book as a \dvr~ and its sample chapters as \pvr's.  We further create virtual books' \dvr's and the randomly selected chapters as \pvr's.  Hence, we hypothesize that the mean distance between the chapters' \pvr~ and the authentic book \dvr~is significantly smaller than that of chapters' \pvr~ and virtual books' \dvr.
We perform a sequence of three experiments, controlling first for the book and chapter lengths, and with each experiment, we relax length restrictions to see the validity of the method for texts of different sizes, adding a sensitivity test. 
Table~\ref{tbl:bv-Results} depicts our results. The results are averaged over 30 real books and 30 virtual books. Each real book contains between 7 to 61 chapters. Virtual books were created with a randomly chosen number of chapters from the pool of real books. Each virtual book contains 19 to 31 chapters.
\begin{table*}[t!]
  \caption{Distance as a measure of authorship: validation and sensitivity}
  \label{tbl:bv-Results}
  \resizebox{\textwidth}{!}{
  \begin{tabular}{ l l c c c}
    \toprule
Experiment & Description & Average & Median & Std.\\
\midrule
 Normalized baseline& Distance of a 1,000 words chapter from a 10-chapter authentic book & 0.306 & 0.304 & 0.057 \\
 & Distance of a 1,000 words chapter from a 10-chapter virtual book & 0.717 & 0.511 & 0.398 \\
 \hline
 Normalized chapters & Distance of a 1,000-words chapter from an authentic book & 0.369 & 0.379 & 0.061 \\
 &  Distance of a 1,000-words chapter from a virtual book & 0.78 & 0.654 & 0.162 \\
\hline
No normalization&    Distance of a chapter from an authentic book & 0.474 & 0.453 & 0.121 \\
 & Distance of a chapter from a virtual book & 0.667 & 0.654 & 0.162 \\
\bottomrule
\end{tabular}}

\end{table*}
The corresponding values of independent two-sided t-tests are the following: Normalized baseline: T(18)=3.2326, p=0.0046; 
Normalized chapters: T(58)=13.004, p=0.0001; 
No Normalization: T(58)=5.228, p=0.0001.
Hence, we can reject the null hypothesis. Hence, the distance of chapters' \pvr~ from their authentic books \dvr~is significantly smaller ($p<0.005$) than that from their distance to the virtual books. This also holds when text length is not normalized.

\subsection{Using \lpa's entities distances to detect that multiple authors wrote a single piece of text}
\label{sec:multiple}
Recall that at the heart of the \lpa~construction method lies the assumption that a domain's long tail distribution is the sum of individual language usage patterns manifested in personal vocabulary distributions. The head of the \dvr~then can be seen as a {\em superposition} of the heads of the entities' vocabularies distributions. As discussed in the Introduction, this is an underlying, sometimes latent,  assumption in many language models that identify topics using prominent terms' frequencies. 

Building on this motivation, we hypothesize that text's vocabulary frequencies written by multiple authors will resemble the \dvr's frequencies, especially in the head, as both vectors result from a superposition of individual terms' usage patterns.  Recall that the head of the \dvr~reflects the typical use of popular terms in the domain. \lpa~identified the personal differences of an individual as their style. Hence, we can assume that when the head of an entity's distribution is very similar to that of the domain, it results from a superposition of several authors' styles. Hence, the hypothesis is that entities' text, when the entities are very close to the domain, is written by multiple individuals. 

How do we show it? To that end, we return to the process we have used before of creating virtual books. Recall that a virtual book is a collection of chapters randomly chosen from other authentic books and collected together as a collection we term 'virtual book'. Thus, taken together, we have a collection of authentic books and virtual books. 
Thus, our domain is the entire Books dataset, and each authentic book and virtual book is an entity.
When constructing the \dvr, we consider only the authentic books (As the \dvr~is constructed in a bag-of-words approach and the virtual books are composed of chapters from the authentic books, both \dvr's, for the virtual books and the authentic books, are identical). 

\begin{figure}[h]
 \centering
 \includegraphics[width=\textwidth]{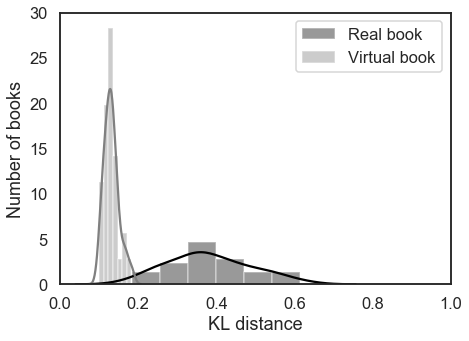}
 \caption{Distances from the \dvr~ for both authentic books and virtual books computed over the Books dataset}
 \label{fig:virtualbooks}
\end{figure}

We then compare the distances of authentic books from the \dvr~ to that of the virtual books. Figure~\ref{fig:virtualbooks} shows the distribution of distances from the \dvr~ for both virtual and authentic books.
We find that the virtual books are both much closer to the \dvr~ and present a much narrower distribution. 
This indicates that distance can be a good indicator of whether a specific text was written by multiple authors and is thus a superposition.

\subsection{Similarity between \pss~as a measure of context-based authorship attribution}
\label{sec:socks}
Here, we define a domain as '19th century-novels' and create a subset of the Books dataset as a domain. The subset contains nine books authored by different authors. Three of the books were authored by Jane Austen, and two books by the Bronte sisters, each a book. The other novels were authored by Lewis Carol, Charles Dickens, Mark Twain, and Arthur Conan Doyle. We create the \dvr~of all nine books and the corresponding \pss~for each novel. We then compute the similarity between the signatures. 

Computing the similarity is not trivial once missing or underused elements are included in the signatures. \pss~ have both weights and signs. A term missing from the entity's vector but frequent in the domain can be highly important in the user's signature, resulting in a high absolute weight, but with a negative sign, indicating that it is missing. When comparing signatures, we would like to take into account the sign. Consider two entities with the same term in a high place in their signature; One uses it more than the typical domain usage, and the other never uses it; This difference should contribute significantly to their distance. Hence, to compare, we add one to all the absolute weights that reference a term with a positive sign in the signature and subtract one from the absolute weight of terms that have a negative sign in the signatures, and then compute the distances using a simple \dln~metric. 

\begin{figure}[!th]
 \centering
 \includegraphics[width=\textwidth]{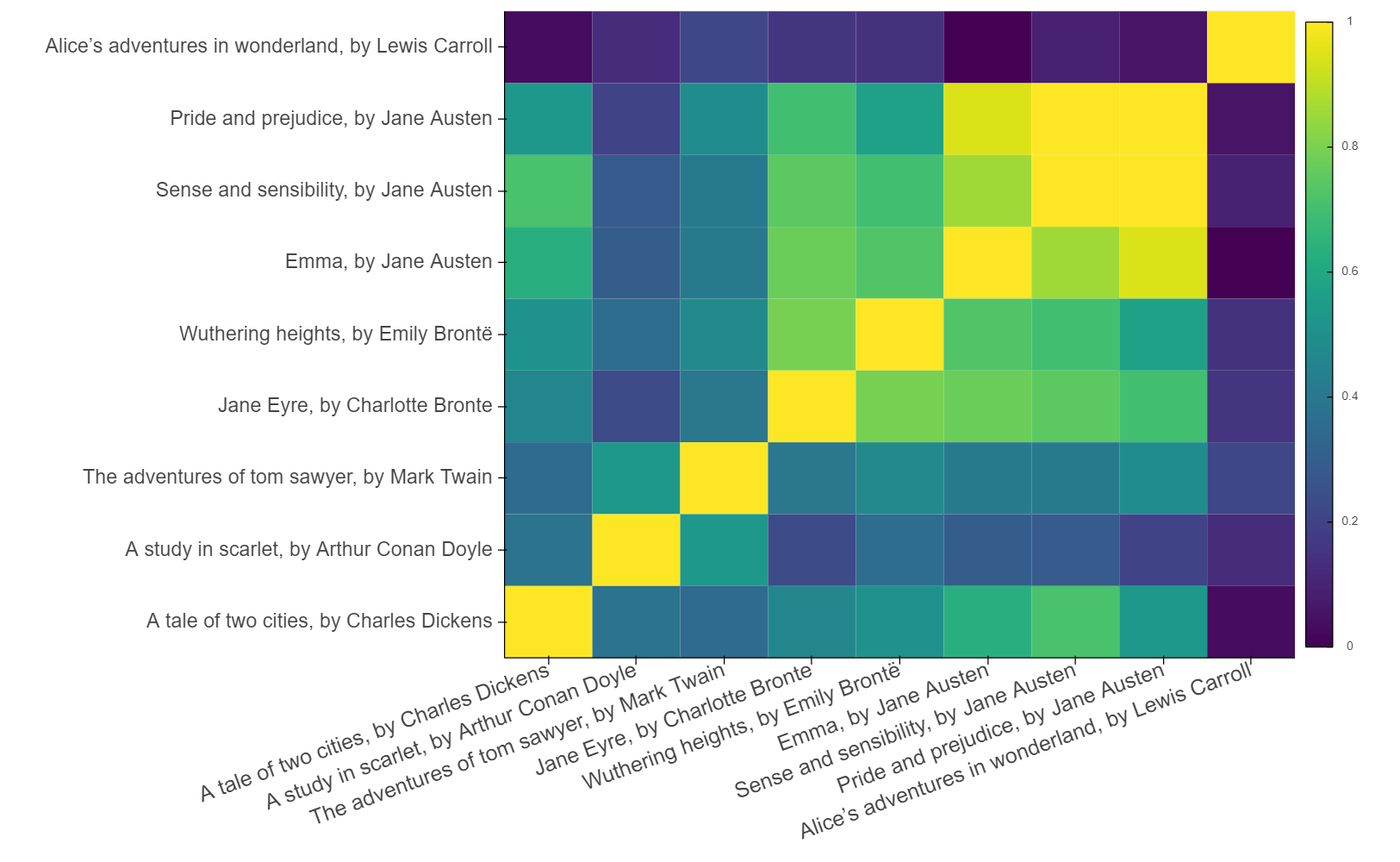}
 \caption{A similarity heatmap (1-distance) between nine 19th century-novels' signatures. }
 \label{fig:heatmap}
\end{figure}
Figure~\ref{fig:heatmap} depicts a similarity heatmap (1-distance) between the novels' signatures. that is, two novels' signatures are very close if the corresponding similarity is one. All three Jane Austen books have similar \pss, with Pride and Prejudice and Sense and Sensibility closer to each other than to Emma. The novels by the sisters Bronte have more similar signatures than any other two novelists; however, their signatures are also relatively close to those of Jane Austen's novels.  Interestingly, the signatures of the novels Sense and Sensibility and A tale of two cities by Dickens are also somewhat close. The most different signature is that of the novel Alice's adventure in wonderland by Lewis Carrol. 

\begin{figure}[!th]
 \centering
 \includegraphics[width=\textwidth]{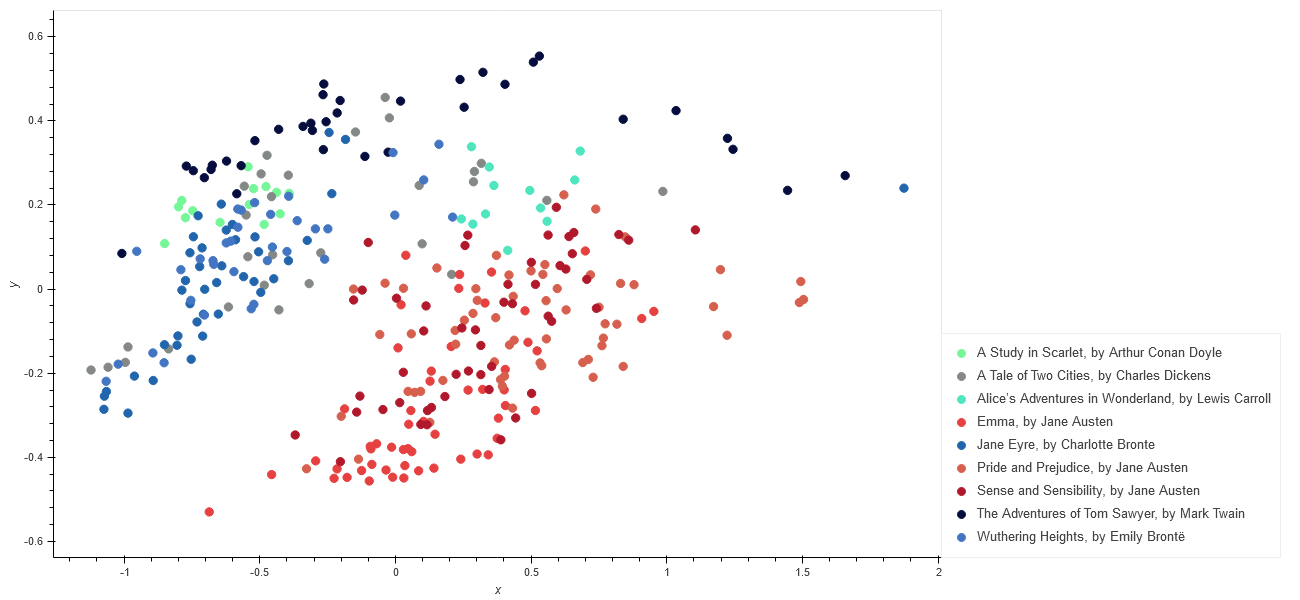}
 \caption{A PCA of the distance matrix table of all 19th-century novels' chapters. The first dimension accounting for 28\% of the variance, the second for 8\%.}
 \label{fig:PCA}
\end{figure}
To identify the differences between personal styles more clearly, we increased the resolution by treating each chapter of the 19th century-novels as an entity. Each chapter out of the 360 considered is a sample of its authors'  writing, and we, therefore, wanted to test to what extent \lpa~ would identify and differentiate individual authors' writings when the chapters are the entities. We created a distance matrix table of all chapters \pss. A PCA of these distances, depicted in Figure~\ref{fig:PCA} accounted for 36\% of the variance in its first three dimensions, with the first dimension accounting for 28\%.  Here, we find that the writings of Jane Austen diverge from the rest, as are the styles of the Bronte sisters and Dickens in a Tale of two cities.  

\begin{figure}[!th]
 \centering
 \includegraphics[width=\textwidth]{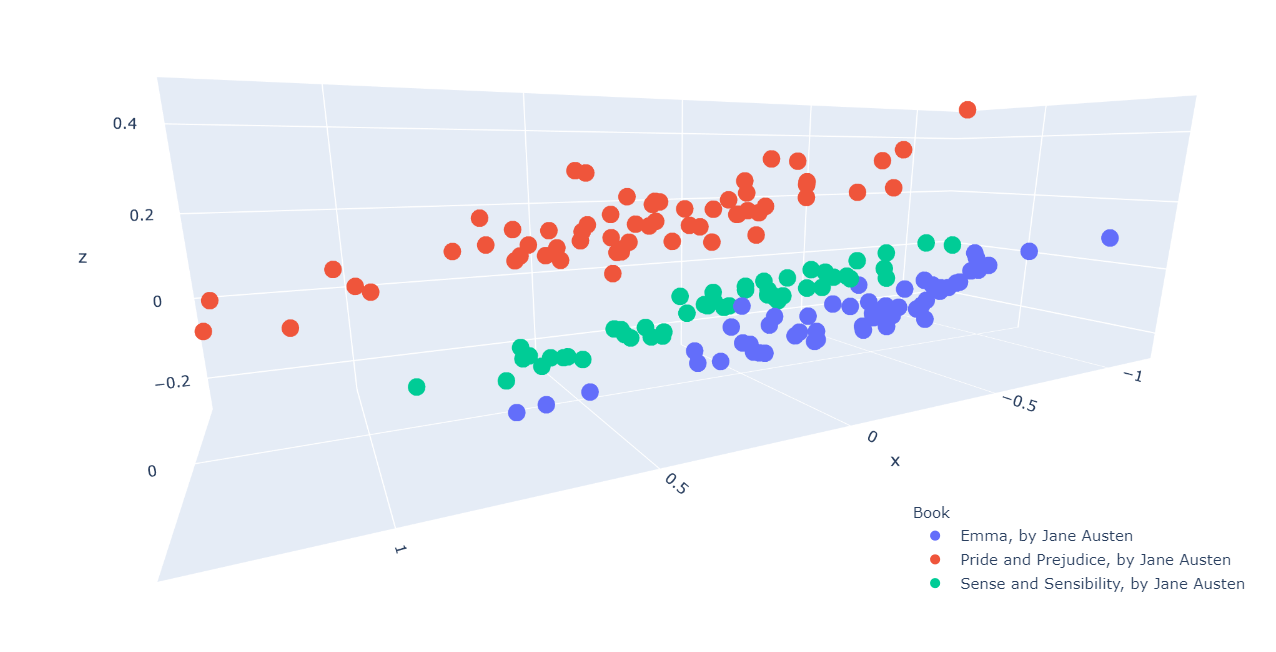}
 \caption{A PCA of the distance matrix table of Jane Austen's three novels' chapters. The first dimension accounts for 25.6\% of the variance, and two more dimensions added together 7\% to a total of 32.5\% of the explained variance}
 \label{fig:austen-pca}
\end{figure}
Jane Austen's novels show high similarity. To further drill down into it, we created a \dvr~and \pss~out of the three novels' chapters and created a corresponding distance matrix. Figure~\ref{fig:austen-pca} depicts the PCA of the distance matrix. The first dimension accounts for 25.6\% of the variance, and two more dimensions added together 7\% to a total of 32.5\% of the explained variance. There is a clear separation between the three novels, with Pride and Prejudice diverging from the other two novels. 

Thus, while \lpa~ captures a personal style and can be used as an authorship-attribution method, its content-based approach enables it to differ between the same author's writings. That is, \lpa~enables a context-based author attribution.

\subsection{Entities required sample size}
\label{sec:size}
To understand the limitations of \lpa, we conduct the following experiment to identify the minimal amount of text needed to identify a signature correctly.  We consider each book as a domain and each chapter as an entity in the domain. The optimal chapters' signatures are obtained when the entire chapter is considered. However, it could be that the prominent terms that set a chapter apart from the domain can be determined using less information. We term the signatures obtained with entire chapters {\em optimal signatures} and conduct a set of experiments to find the minimal amount of information needed to obtain a close-to-optimal signature. 

For each book $i \in$ the Books dataset, we find the average length of a chapter in a book, $X_i$, and vary a sample length, $j$  in the range $L={0.1X, 0.2X, \ldots 1.2X}$. As $X_i$ is an average chapter length, some chapters will be shorter than $X_i$ and some longer. For each sample length $j \in L$, we create a \dvr~and signatures for book $i_j$, that is, the sampled book $i_j$ that consists of chapters of length $j$. For example,  assume $j=0.1X$.  Each entity (chapter) is a frequency vector formed from the chapter sampled with the first 10\% terms (10\% of the average chapter length). The \dvr~is the aggregated frequency vector of the combined entities, and the signatures consist of the terms most distancing the entities (sampled chapters' vectors) from the \dvr.
We then compute for each entity's signature its $L_1$ distance from the corresponding chapter's {\em optimal signature}.  The number of terms in each book corresponds to the number of nouns and adjectives. 

\begin{figure}[!th]
 \centering
 \includegraphics[width=\textwidth]{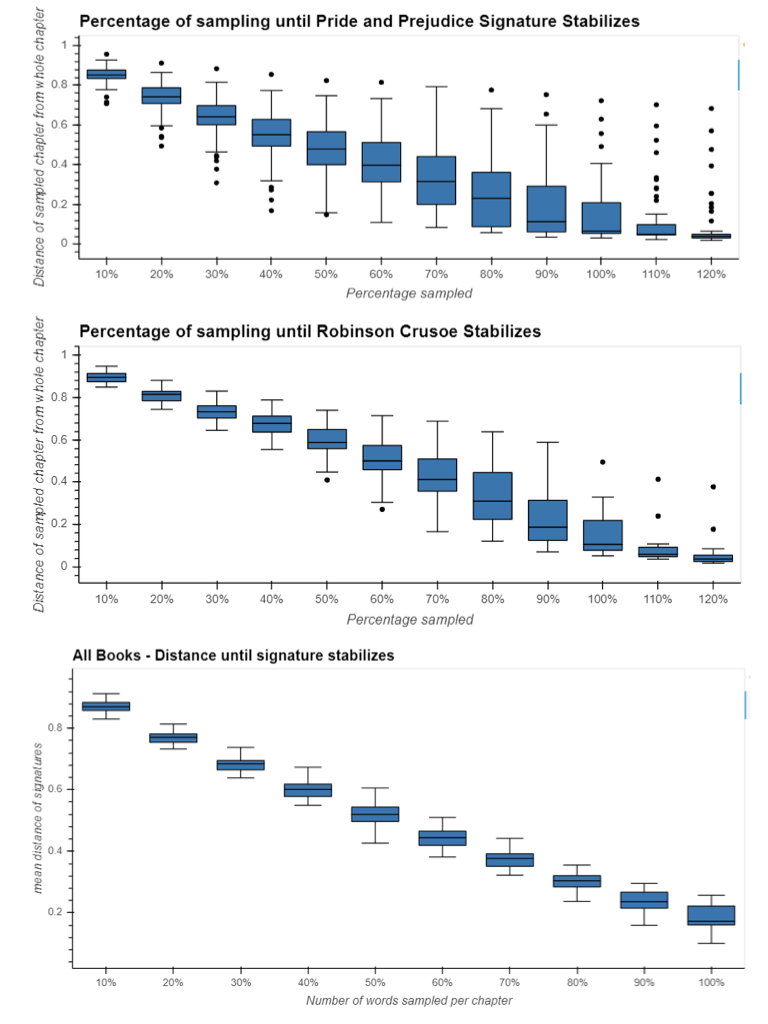}
 \caption{Sample size that yields a signature close to the one received with full information. Upper panel: Pride and Prejudice; Middle panel: Robinson Crusoe; Lower panel: Average over the entire dataset. Each Box plot corresponds to the average over the 30 books, a total of 943 chapters.}
 \label{fig:stabilize}
\end{figure}
Figure~\ref{fig:stabilize} shows the Box plot of the distances of the sampled chapters per the Novels Pride and Prejudice, Robinson Crusoe, and for the entire Books dataset. The novel Pride and prejudice consists of 61 chapters with an average chapter length of 780 terms. The life and adventures of Robinson Crusoe contains 20 chapters with an average length of 2060 terms. The upper two panels show the average distance between the sampled chapters' signatures and the {\em optimal signatures} for each sample length. The signatures are close to the optimal signatures for both books when we get to a sampling that corresponds to 0.8 of an average chapter's length. The lower panel depicts the average result over all the books, i.e., an average of averages. There are thirty books and a total of 943 chapters. In each experiment, we sample the chapters ten times on different lengths and compute the distances' average for all sampled chapters. Thus, each point is an average of 30 averages, each taken roughly over 31. 4 chapters.  Books differ in their number of chapters and chapters' length. However, on average, the signatures become close enough to the optimal when we take 0.8 of the average chapter length. 
In summary, the signatures are more accurate the more information is used.

\section{\lpa~in a social media domain}
We continue to evaluate \lpa's performance over a large-scale real-world social media dataset in the movies' domain. Here, we use the data from the IMDb review-platform site. It offers an opportunity to evaluate the performance of \lpa~ on a domain-based online free short text, and in particular, to identify two types of impersonation.  

We identify two types of inauthentic users in a social media domain. One is {\em multiple virtual personas} (recently referred to as Sockpuppets when the accounts are used for various goals ~\citep{kumar2017army}), which are multiple user accounts operated by the same person.  The other is what we term {\em front-users}, a single virtual persona operated by multiple people. \lpa~ lends itself to find front-users. Its definition comes from finding the \dvr~as the superposition of different styles. Hence, as we have shown in  Section~\ref{sec:multiple}, accounts that are also a superposition of several styles will be typically closer and similar to the \dvr. 

We start by evaluating the capabilities of \lpa~ to identify the first type of impersonation, two user-accounts with the same author. Recall that \lpa~ is an unsupervised method that creates vector presentations (signatures) for entities (users) in a domain,  unlike current learning-based methods for authorship attribution in the wild~\citep{abbasi2008writeprints,koppel2014determining}. 
We randomly choose 1000 users, each having between $4000$ to $7300$ words in all of their reviews combined from IMDb. The text of 10 randomly chosen users out of 1000 is halved, and two users are created from it, resulting in 1010 users. In the baseline method, we create the \tf~ vectors for all 1010 users and compute the cosine similarity between all the vectors. 

In our method, we create a DVR from the text of the 1010 users and compute their signatures. We then compare the signatures of all the users to each other ($510050$ comparisons). In Section~\ref{sec:socks}, we have established that texts written by the same author have relatively closer signatures than to other arbitrary texts and described how we compute the distance taking into account that the same term can be either missing or over-used by two users. Recall that in \lpa, signatures have both weights and signs. A popular term missing from the user's vocabulary can be highly important in the user's signature, resulting in a high absolute weight, but with a negative sign, indicating that it is missing. We use for comparisons the algorithm described in Section~\ref{sec:socks}. To determine what \textit{relatively close} means, we define a threshold as follows. We compute the average distance between all pairs, $\sigma$, with $\mu$ corresponding to the standard deviation. We then set the threshold to $\sigma+r\cdot\mu$, where $r=1 \ldots 4$ in the different experiments.
We determine that two user accounts have the same author if the distance between the accounts' signatures is lower than the threshold.  We can expect that a lower threshold increases the precision but decreases the recall. 
 
\begin{table}[h!]
\centering
  \caption{Authorship Attribution: LPA vs. baseline}
  \label{tbl:valid}
  \begin{tabular}{c c c c c c c}
    \toprule
Threshold&   \multicolumn{3}{c}{Baseline} & \multicolumn{3}{c}{Our method}\\
& Prec. & Recall & F1 & Prec. & Recall & F1\\
\midrule
Avg - 1 std. & 0.96 & 0.5&0.65&0.87&1&\textbf{0.93}\\
Avg - 2 std. &0.98&0.68&0.8&0.96&1&\textbf{0.98}\\
Avg - 3 std. &0.98&0.77&0.86&0.99&1&\textbf{0.99}\\
Avg - 4 std. &0.99&0.83&0.9&0.99&1&\textbf{0.99}\\
  \bottomrule
\end{tabular}
\label{tb:eval-sim}
\end{table}
Table~\ref{tb:eval-sim} details the results of the comparisons for both the baseline method and our method for the four threshold values. Our method outperforms the baseline in all cases. It always determines correctly the users that their text was halved as being the same author, with very low false-positive values, i.e., users that are mistakenly considered the same author. 

\subsection{Finding multiple personas (Sockpuppets) in big-data}
Multiple virtual personas are several accounts operated by the same author, also termed Sockpuppets. As such, we expect them to have relatively close signatures in the domain. Continuing from above, we employ the method here over the entire Social Media dataset to find multiple users that belong to the same author on the Social media dataset. Our dataset contains $N=3969$ users, yielding $N^2=7,874,496$ comparisons. Our method computes the $L_1$ distance between them, and hence computation-wise takes $O(N^2)$ time complexity over a memory requirement of $O(k \cdot N)$, $k=500$ terms. Table~\ref{tbl:socks} details the number and percent of users our method identifies as multiple personas - i.e., their signatures are close to each other. We compute signatures as close using four different thresholds: one std, two std, and up to four std. 's below the average.  
\begin{table}[h!]
\centering
  \caption{Social media dataset Sockpuppets results}
  \label{tbl:socks}
  \begin{tabular}{ c c c}
    \toprule
Threshold&  User pairs suspected as Sockpuppets & Percent \\
\midrule
Avg - 1 std. & 1,180,261 & 14.988\%\\
Avg - 2 std. & 336,173 & 4.269\% \\
Avg - 3 std. & 86,975& 1.104\%\\
Avg - 4 std. & 21,996& 0.279\%\\	
  \bottomrule
\end{tabular}
\end{table}
We conducted manual verification for a randomly chosen number of suspected users from this group (Avg - 4 std.). We found corroborating evidence from their personal information that the same person operates the accounts. For details and the full list of sockpuppets see \url{https://www.scan.haifa.ac.il/post/domain-based-latent-personal-analysis-lpa-and-its-uses}. 

\subsection{Finding front-users in big-data}
The second kind of impersonation we are trying to identify is front-users, one virtual persona operated by multiple people. 

\omitit{
Each of these people would have a unique writing style and, theoretically, a unique weight-term vector. We have no access to these hypothetical vectors, but we know that the front-user's actual term-weight vector will be a weighted average of these vectors. Such a weighted average would tend to decrease personal idiosyncrasies of single writers and increase shared habits. In other words, imagine five people are writing under the same pseudonym, and imagine each contributes roughly an equal share of the text. While each of them would have a personal style and would tend to use specific words more frequently, such differences would average out when considering the entire text as a whole. This is the exact same process, albeit on a much smaller scale, as when constructing the \dvr. We, therefore, expect a front-user's vector to be relatively close to the \dvr.
}

\begin{figure}[h]
 \centering
 \includegraphics[width=.98\textwidth]{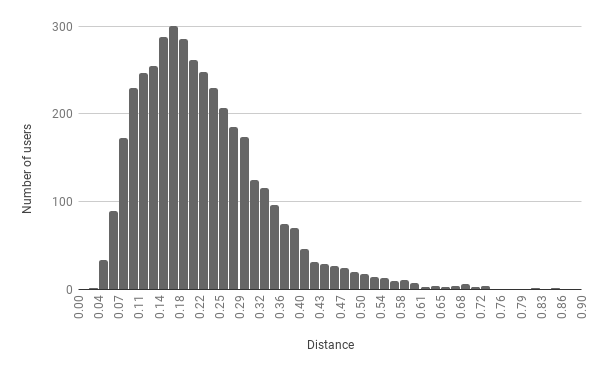}
 \caption{Distribution of Social media users' distances from the \dvr}
 \label{fig:IMDBfrontusers}
\end{figure}

Similar to our finding in Section~\ref{sec:multiple}, we find here front-users in a real-life dataset, our Social media dataset.
To identify front-users, we compare the distance of the $3970$ users to the \dvr. We expect a front-user's vector to be relatively close to the \dvr. We define the threshold as follows. We calculate all users' distance from the \dvr and find the average and standard deviation. We then define a lower threshold, e.g., a standard deviation under the mean. Any author whose vector's distance is lower than this threshold is considered a possible front-user. We find that the average distance is 0.223 (Median: 0.203), with a standard deviation of 0.114 (the distance distribution is right-skewed). We consider users whose distance is not more than one standard deviation below the average, that is, their distance is below $0.109$, as suspected front-users. In our sample, $532$ users fit this criterion.  Figure~\ref{fig:IMDBfrontusers} shows the distribution of the users' distances. 

To find evidence supporting our findings, we explore these users' temporal rates utilizing visual temporal analysis and temporal measures. 
We are particularly interested in two parameters: i) the number of active days, and ii) maximum posts per day. 
To find a non-normal activity pattern, we disregard periods of inactivity. This allows us to depart from a timeline view and present a view that concentrates on periods of activity. We can thus easily detect peaks of extreme activity, which indicate inauthentic usage. Imagine, for example, that we find on the Social media dataset a user who has not been active for two months and then posted 60 reviews in one day. While the average would be one review per day, which is not suspicious in itself, it is evident that the user had not watched and reviewed 60 movies in one day.
Alternatively, imagine a user who has been active entirely consistently every day, posting precisely one review every day. That, too, is inconsistent with normal human behavior, characterized by bursts of activity~\citep{barabasi2005origin}.  We find these two patterns visually with Activemap~\citep{ben2018activemap}, a Treemap visualization designed to highlight two types of outliers on social media: contributing users with a substantial amount of contributions or with extreme activity peaks. We Visualize the $532$ suspected front-users with Activemap,   allowing us to quickly identify that many show a larger than normal activity volume either over the entire period or during specific days.  Considering these are reviews for films, it is improbable, for example, for one user to write fifty such reviews during a single day, as is the case for some.

We continue with the temporal validation to quantify deviations from normal behavior patterns. We are particularly interested in two parameters: i) the number of active days, and ii) maximum posts per day. 
As we are interested only in days of activity, we disregard periods of inactivity. This allows us to depart from a timeline view and present a view that concentrates on periods of activity. We can thus easily detect peaks of extreme activity, which indicate inauthentic usage. Imagine, for example, that we find on the Social media dataset a user who has not been active for two months and then posted 60 reviews in one day. While the average would be one review per day, which is not suspicious in itself, it is clear that the user had not watched and reviewed 60 movies in one day.
Alternatively, imagine a user who has been active consistently every day, posting precisely one review every day. That is inconsistent with normal human behavior, which is characterized by bursts of activity~\citep{barabasi2005origin}.  
Therefore, we are interested in two kinds of users - those who consistently post a large number of reviews or those with extreme peak activity.

We visualize the suspected front-users activity utilizing Activemap~\citep{ben2018activemap}, a Treemap visualization designed to highlight two types of outliers on social media: contributing users with a substantial amount of contributions or with extreme activity peaks. This would translate to users that are either consistently posting a large number of reviews or with an extreme peak of activity. 
\begin{figure}[h]
 \centering
 \includegraphics[width=\textwidth]{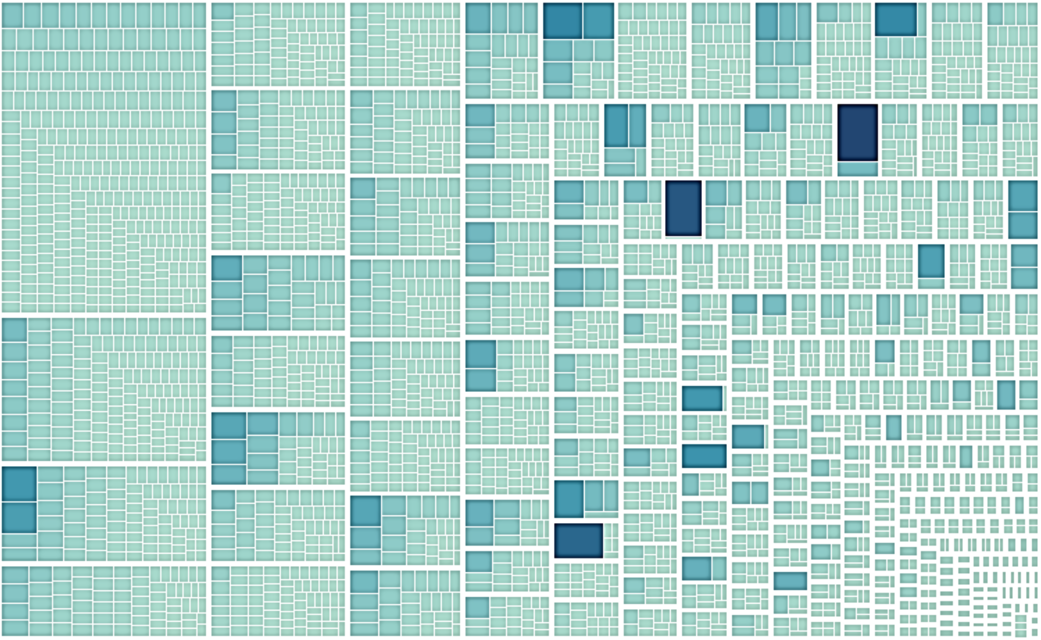}
 \caption{ActiveMap of the 532 Social media users identified as potential front-users}
 \label{fig:treemap}
\end{figure}
Figure~\ref{fig:treemap} depicts the temporal activity of the above found $532$ Social media users as suspect front users, with a distance of less than $1.09$ from the \dvr. The visualization includes only days in which users are active and post at least five different posts. Users are separated with thick white lines, and their days of activity are separated with internal thin white lines. Each Activemap branch of the tree denotes a user, and each leaf is a single day. The leaves show a double mapping of both size and color -  days with high activity are larger and darker. Users are arranged in order of the number of active days, with users with the least active days at the bottom-right corner and users with most active days in the upper left.
This visualization allows us to quickly identify that many of these users show either a larger than normal activity volume over the entire period or during specific days.  Taking into account these are reviews for films, it is highly unlikely, for example, for one user to write fifty such reviews during a single day, as is the case for some.

In summary, we see that \lpa~can easily find suspected front-users in a social media domain. 

\section{Related Work}
Power law distributions are prevalent in nature, and more so in empirical data~\citep{newman2005power}. Power-law distributions were found in a wide-area of situations, among them allometric scaling in animals~\citep{west1997general}, number of citations received by scientific papers~\citep{price1965networks}, frequency of personal names~\citep{zanette2001vertical}, and population of US cities as recorded in 2000~\citep{newman2005power}. In a power-law distribution, extreme values are more likely than they would be in a Gaussian distribution, for example. While the latter distributions are centered around a typical value, long-tail distributions do not peak at an expected value and may vary over many orders of magnitudes. 

In languages, it was~\cite{zipf1949human} who demonstrated that this type of decaying relationship exists between a rank of a word and its frequency.  We speak to convey meaning, and indeed, "word meaning is a substantial determinant of frequency, and it is perhaps intuitively the best causal force in shaping frequency"~\citep{piantadosi2014zipf}. In that sense, the Zipf distribution can be seen as a 'need' distribution, related to how often we need to convey each meaning. Interestingly, popular words tend to have multiple meanings~\citep{i2003least}, yet they are also easier to produce~\citep{brown1966tip,traxler2011handbook}. Zipf's Least effort explanation relates to a lexical tradeoff between the speaker's utility of resorting to the easy to produce, popular words and the hearer utility, which requires clarity and hence longer sentences~\citep{zipf1949human,ferrer2018origins,Hahn2347}. 

Vast literature supports the assumption we make here, that in each domain, there is a subset of the frequent, popular words. The Swadesh list contains universal, cultural independent available words, including "I", "one", "many", and "water", for example. An examination of frequencies of words on the lists across 17 languages found a high correlation, suggesting that these words are predictable from their meaning~\citep{calude2011we}. Similar findings were obtained for numbers~\citep{piantadosi2014zipf} and taboo words~\citep{piantadosi2014zipf}. The same words may carry different meanings in different topics, therefore having various frequencies per various topics~\citep{griffiths2002probabilistic,hofmann2013probabilistic,blei2003latent}. Hence, our method applies within a domain where words have a specific meaning and a corresponding frequency and popularity. 

In this work, we showed how \lpa~ creates domain-based vector representation (signature) and distance for each entity in the domain. We demonstrated that both could be used for explorations in domains and aiding in user modeling problems. Specifically, we show that it lends itself to authorship attribution and impersonation detection problems. As \lpa's entities' attributes are slim and compact, we found it especially suitable for exploring online social media sites' impersonations. 

\subsection{Social media}
The prevalent anonymity within the cyber world enables unethical behaviors such as impersonation, the utilization of multiple online identities, and the spread of false information. Malicious behavior in crowdsourcing platforms is widespread and used for both fun and profit~\citep{wang2012serf}.  The problem of online anonymity in the cyber world is addressed by employing authorship analysis and attribution techniques ~\citep{zheng2006framework,abbasi2008writeprints,stein2011intrinsic,iqbal2013unified}. For example,~\cite{barbon2017authorship} suggest to extract baseline style features for an account and compare the style of new content in the account to determine if the account was compromised.    However, social media text introduces specific challenges as it is often short, unstructured, and informal~\citep{krippendorff2018content}.  

Our primary interest is in identifying impersonations of the following types:  authors that maintain multiple accounts, also termed sockpuppets~\citep{kumar2017army}, and front users,  a term for single accounts that are each maintained by several authors. While the latter has gained little attention, the former, i.e., detecting sockpuppets, has attracted much attention, especially from the authorship attribution community.

\subsubsection{Social media authorship attribution}
Authorship attribution is the science of inferring characteristics of the author from the characteristics of documents written by that author~\citep{juola2008authorship}. Authorship attribution methods famously brought quandaries to a conclusion:  they distinguished the different authors of the Federalist papers \citep{Mosteller1964} and established that the Chinese history book 'Dream of the Red Chamber', originally thought to be written by a single author,  was the product of a collaborative work~\citep{hu2014multiple}. 

Authorship analysis, rooted in stylometry, is the process of examining the characteristics of a piece of writing to conclude its authorship.   Authorship identification determines the likelihood that a particular author produced a piece of writing by examining other writings by that author~\citep{zheng2006framework}.  
Until the late 1990s, authorship analysis research was dominated by attempts to define stylometry writing style features.~\cite{Holmes1998} explored the use of text length and vocabulary richness to identify a unique writing style per author.~\cite{Burrows1987}  found that the 100 most frequent words can represent an author's style.  

The modern analysis methods turn to classifiers. 
According to a predefined feature-set, classification-based authorship attribution methods learn each author's style and determine the author of a new text accordingly. 
~\cite{narayanan2012feasibility} show that traditional classification methods for authorship analysis do not perform well over large real-world corpora of small texts. For example, Linear Discriminant Analysis does not perform well due to the data's sparsity, while simple classifiers perform better. They further demonstrate that it is challenging to scale classification beyond a few hundred authors. However, this is the setting in 'the wild' - large real-life corpora with a large open candidate set~\citep{koppel2011authorship}. They further identify that the large candidate-set introduces a  scalability problem.   A single classifier cannot be trained to classify to $N>>1$ (10,000 authors in their case) classes, even with a small feature set.  A one vs. all solution with $N$ binary classifiers is also not feasible. Instead, they repeatedly choose a fraction of the feature-set and find top matching candidates using cosine similarity. The chosen candidate, most times as a top match, is selected as the author of the anonymous text.  ~\cite{koppel2014determining} explore the use of similarity methods. They claim that naive similarity approaches do not work well when the text is short and the candidate set is large. Instead, they suggest to add impostors from the same domain and repeat the similarity test over a different subset of the features. The writer that scores highest on the majority of the tests is chosen as the author. \cite{rocha2016authorship} review authorship analysis methods for detecting online deceptions and impersonations in open source coding authorship,  social networks, and social media, and discuss the limitations of classification-based methods.  Current state-of-the-art works consider, among other attributes, character-grams, and used symbols such as preferred punctuation marks or emojis~\citep{koppel2009computational,schwartz2013authorship,shrestha2017convolutional,neal2018surveying}. 

\lpa~creates, for each entity, a distance and a slim vector representation with respect to a domain, and can thus be applied to distinguish between thousands of authors. We conducted an all vs. all experiment for over 4000 authors, yielding 16 million comparisons, to detect authors of multiple social media accounts. However, while the works above considered {\em any} online text and included an extensive set of supporting features, we merely work with used text frequencies and require that all text samples relate to a specific domain. 
In this sense, our work is closer to Plagiarism, which measures the distance between texts to establish whether a single author has written them, but makes no attempt to identify or characterize the author~\citep{clough2000plagiarism,vani2018unmasking}.  Another field in which prominent works utilize similar methods is text categorization. \cite{bigi2003using} used a back-off probability model to categorize texts in large corpora by measuring the KLD distance between the probability distribution of a document and the probability distribution of each category. Latent Semantic Indexing (LSI) analysis algorithm~\citep{deerwester1990indexing} is a document indexing method to optimize search engines. The method considers synonyms that do not appear in the text yet might appear in other texts. LSI aggregates all the synonyms under the same term with the idea of reducing dimensions in the search process. In our method, we do not aggregate synonyms, as we see the use of a specific term and not another as a signature usage.

The problem of high dimensionality in social media was also addressed by~\cite{zhang2014authorship}. They designed a framework that analyzes the text to find features and then reduces dimensions before performing the classification.  They found that simple distance cosine-metric over a \tf~vector representations of the text can attribute over 56\% of the samples to the correct author. They were able to improve the result by utilizing meta-learning methods. In this work, we do not use a large set of different features to describe an author.  We further showed that \lpa outperforms \tf~in identifying authorship of multiple social-media accounts. 
\subsection{Temporal analysis}
One of the application suggested in our work is identifying front-users, a social media account operated by multiple authors. We then corroborate our findings by looking at their online temporal activity. In the following, we give a short review of online temporal activity patterns and anomalies. 
\subsubsection{Normal activity patterns}
Users are not expected to be consistently active. \cite{BA99} found increasing evidence that a wide range of human activities shows bursts of extensive activity, separated by long periods of inactivity.  Hence, we will consider suspicious a user with a very consistent activity pattern. \cite{ferraz2015rsc} characterized the distribution of postings inter-arrival times of users postings and find four categories of behaviors that show repetitive patterns. They suggested a Rest-Sleep-and-Comment (RSC) generative model that accommodates these categories to model users' online behavior on social media sites. \cite{viswanath2014towards} studied the behavior of black-market activity in social media networks (purchase of Facebook likes). Their study found that the granularity of a single day gave the best results. While they considered the number of likes per day, they also remarked that the same could be applied to any other user behavior. We use the number of published reviews per day as our user activity baseline for our temporal method.
\subsubsection{Analytic visualization}
Social media data visualization has become a conventional method to analyze and summarize information ~\citep{schreck2013visual}. Visualizations exist for information flow  ~\citep{chen2016d}, sentiment divergence ~\citep{cao2015socialhelix} and early flood detection ~\citep{johnson1991tree}. Visual techniques are used for anomaly detection in a variety of settings: Internet security utilizing Border Gateway Protocol data ~\citep{steiger2014computer}, GPS data ~\citep{kietzmann2011social}, Local events on Twitter and Sensor networks ~\citep{snijders2001statistical}.
Here, we present and employ a tree-map visualization specifically designed to highlight the characteristics we are interested in - consistency and intensity of posting periods.

\section{Discussion}
In this work, we have presented \lpa, a method for finding latent personal signatures of entities in a domain.  In a domain comprised of a collection of entities' elements, \lpa~ creates two representations for each entity:  its relative distance from the domain and its relative signature. The relative signature is a vector representation determining which elements most separate the entity from the domain. Created initially for text, we considered here specific textual domains, in which each entity is a text sample (a book, a chapter, or a social media user account), and the elements are the terms. We explore the different characteristics and uses of the entities' distances and signatures for context-based authorship attribution and two types of impersonations in a social media domain: authoring different user accounts while maintaining different online personas for each, and front-users, which are accounts maintained by several authors pretending to be a single author. 

\begin{figure}[!th]
 \centering
 \includegraphics[width=1.1\textwidth]{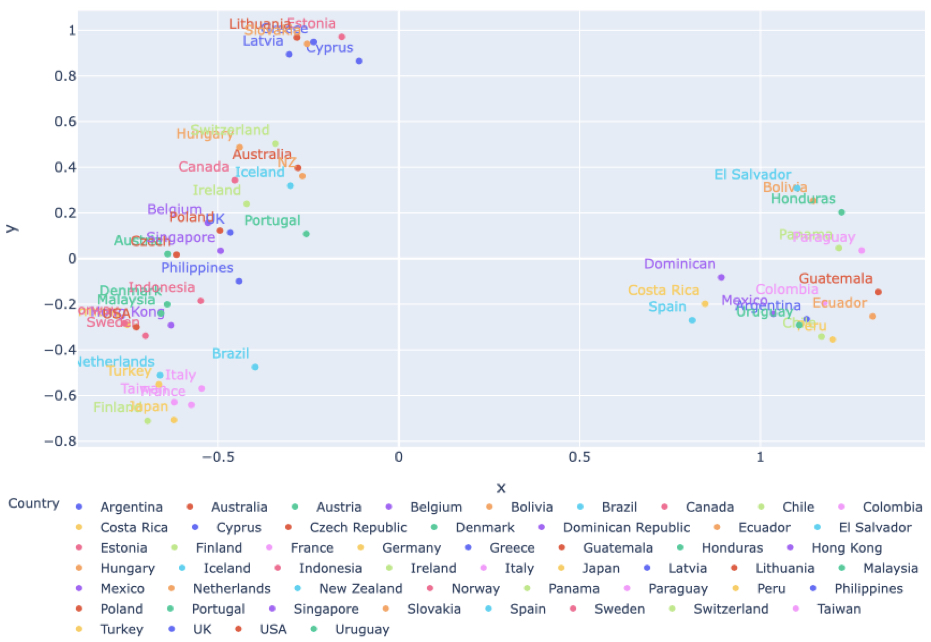}
 \caption{A PCA of the distance matrix table of Countries Spotify music \lpa-signatures. The first dimension accounts for 47.7\% of the variance, and the second for 16.8\%.}
 \label{fig:music-pca}
\end{figure}

We demonstrated that \lpa~creates signatures that are meaningful in several domains. In Section~\ref{sec:examples} we demonstrated the construction of \lpa~distances and signatures for countries Spotify songs listening habits. Countries with a smaller distance to the domain had similar listening habits to that of the domain (mostly listened to popular songs). Countries more distant from the domain tend to listen more to local music (while still listening to popular music). Two similarly distant countries may have very different listening habits, as is determined by their domain-relative signatures. To illustrate which countries have similar listening habits, we computed the distances between all countries' signatures. Figure~\ref{fig:music-pca} shows the PCA computed over the resulted distance matrix. The first dimension accounts for 47.7\% of the variance, and the second for 16.8\%. Some of the results are intuitive. Spanish-speaking countries like Spain, Guatemala, and Mexico are closer together in the lower right corner. Paraguay and Panama are very close together while being relatively close to the Spanish cluster. On the periphery of this cluster are El Salvador, Bolivia, and Honduras. Interestingly, Cyprus and Greece are very close to Estonia, Lithuania, Latvia, and Slovakia. Another intriguing cluster in the middle left consists of Denmark, Sweden, the USA, Germany, Indonesia, Malaysia, and Hong Kong. 

\lpa~ has several limitations. As computed in Section~\ref{sec:size}, the sample size required to preserve a book's chapter's distance from a domain is of the order of the size of the average book chapter's length. \lpa~also has limitations when used in social media and short text. 
Short-text written online contains typical typos and symbols. Therefore, the state-of-the-art methods for authorship attribution 'in the wild'~\citep{Koppel2013} are the ones identifying these styling differences~\citep{schwartz2013authorship,shrestha2017convolutional}. \lpa, on the other hand, offers an unsupervised yet quick and practical method for determining vector representations for users, and is not equipped to work with unseen text. Instead, \lpa-signature representations may be used as additional features in learning-based authorship attribution methods. 

\lpa~Python code is available from~\cite{alon2021using} in \url{https://github.com/ScanLab-ossi/LPA}. 
\section{Conclusions}
We have demonstrated a method for finding how users differ in their language usage from a domain. We considered four different distance metrics for establishing this difference and, based on several defined criteria, found  \kl~the best suited. Building on this finding, we devised personal signatures for users within a domain, termed \lpa, and determined users' distance from the domain. 
\lpa~ finds a domain-related distance and signature for users, or entities, in a domain. We showed that these could be used for unsupervised exploration within a domain for authorship attribution and impersonation detection.  In social media, \lpa~can be used for identifying sockpuppets and front-user accounts. 
In future work, we intend to explore the use of \pss~as features in learning-based attribution tasks and user modeling in recommender systems. 
\section*{Acknowledge}
The authors would like to thank Tom Atkins and Uri Alon for their help. We would also like to thank David Bodoff and Einat Minkov for interesting discussions and helpful remarks. 
\section*{Declarations}
\textbf{Funding: } The research was partially funded by the Magnet Infomedia Consortium, Israel. 
\textbf{Conflict of interest: }The authors declare that they have no conflict of interest.
\textbf{Availability of code, data and material: } Code and data are available from \url{http://scan.haifa.ac.il/data}. 
\bibliographystyle{spbasic} 

\appendix
\newpage
\section*{Appendix A - Distance Metrics}
\label{sec:apndx-a}
A distance metric $d$ on a set $X$ is a function $d:X\times X\rightarrow [0,\infty) $.  
I.e., it receives two elements in the set and gives the distance between them as a real, non-negative number. To be a true metric, such a function needs to fulfill the following four criteria:
\begin{description}
\label{sec:criteria}
    \item \textbf{Non-negativity: } 
$ d(x,y)\geq 0$.
The distance between any two elements is greater or equal to zero.
  \item \textbf{symmetry: }
        $d(x,y)=d(y,x)$.
      Distance is independent of starting point - for any two elements x,y in the set, the distance from x to y is the same as the distance from y to x.
    \item \textbf{Identity of indiscernibles: } 
  $ d(x,y)=0\leftrightarrow x=y $.
        Two elements have a zero distance from each other if and only if they are the same element.
        \item \textbf{Triangle inequality: }
$d(x,z)\leq d(x,y)+d(y,z)$. 
The shortest path is a direct one - given two elements, x,z in the set, it is never shorter to go through a third one, y.
    \end{description}

We show here that each of our selected metrics, as described in Section~\ref{sec:4dist}, is a distance metric by the definition given in Section~\ref{sec:criteria}. 
Specifically, we show that the discussed distance metrics in~~\ref{sec:4dist} all satisfy the following three properties: Non negativity, Symmetry and Identity of indiscernibles. 

\subsection*{RBD distance metric}

\textbf{Non negativity}:  We first notice that $\sum_{d=1}^\infty p^{d-1}$ is the sum of the geometric progression $p^{d-1}$ and is therefore equal to $\frac{1}{1-p}$. since $A_d \leq 1$, we get $$\sum_{d=1}^\infty p^{d-1}\cdot A_d\leq \frac{1}{1-p}$$ and therefore $$(1-p)\sum_{d=1}^\infty p^{d-1}\cdot A_d\leq (1-p)\frac{1}{1-p} = 1$$  
As $p$ is in the range $[0,1]$ and $A_d$ is non-negative we also get $$(1-p)\sum_{d=1}^\infty p^{d-1}\cdot A_d\geq 0$$ Hence RBO is always in the range $[0,1]$. As $RBD=1-RBO$ it is also in that range. 

\textbf{Symmetry}: For two lists, $V_1,V_2$, $A_d$ is defined by the intersection of the lists over the first $d$ terms. This is a symmetrical property - the intersection of $S$ with $T$ is the same as the intersection of $T$ with $S$. $p$ is an independent parameter, and we therefore have 
$RBO(V_1,V_2,p)=RBO(V_2,V_1,p)$, and hence this also holds for RBD.

\textbf{Identity of indiscernibles}: 
Let us consider $V_1=V_2$, that is for every term $d$, $V_1(d),V_2(d)$. Then $\forall d, X_d=d$ and $A_d=1$. We then have $$\sum_{d=1}^\infty p^{d-1}\cdot A_d=\sum_{d=1}^\infty p^{d-1}$$As before, this is the sum of the geometric progression $p^{d-1}$ and is equal  $\frac{1}{1-p}$. Therefore, for $V_1=V_2$ we get $RBO(V_1,V_2,p)=1$ and $RBD(V_1,V_2,p)=0$. \\
Now, consider two distinct lists. that is, for some term d, $V_1(d)\neq V_2(d)$. For that term we have $X_d<d$ and $A_d<1$.
We therefore have $$\sum_{d=1}^\infty p^{d-1}\cdot A_d<\frac{1}{1-p}$$ and therefore $RBO(V_1,V_2,p)<1$ and $RBD(V_1,V_2,p)>0$.

\subsection*{Cosine Similarity}
\omitit{
\textbf{Non negativity}: First we would like to show that $\frac{V_1\cdot V_2}{\| V_1 \| \times \| V_2 \|}$ is in the range $[0,1]$.  We remember that the name cosine similarity stems from the fact that if we consider $V_1, V_2$ to be vectors in the $\R^n$ Euclidean space than they have some angle $\theta$ between them and furthermore $cos\theta = \frac{V_1\cdot V_2}{\| V_1 \| \times \| V_2 \|}$. \\
**** The following may be unnecessary \\
We can show this by considering the triangle created by the three vectors $V_1, V_2, (V_1-V_2)$. By the law of cosines we have \\ $\|V_1-V_2\|=\|V_1\|+\|V_2\|-2\|V_1\|\|V_2\|cos\theta$. \\
We now remember that the dot product is distributive and that for a vector $V$ we have $V\cdot V=\|V\|$, and therefore \\
$\|V_1-V_2\|=(V_1-V_2)\cdot (V_1-V_2)=(V_1\cdot V_1)-2(V_1\cdot V_2)+(V_2\cdot V_2)=\|V_1\|+\|V_2\| 
-2(V_1\cdot V_2)$\\
We thus have two different expressions for $\|V_1-V_2\|$. We equate them to get  \\
$\|V_1\|+\|V_2\|-2\|V_1\|\|V_2\|cos\theta=\|V_1\|+\|V_2\|-2(V_1\cdot V_2)$ and so\\
$2\|V_1\|\|V_2\|cos\theta=2(V_1\cdot V_2)$ and finally \\
$cos\theta = \frac{V_1\cdot V_2}{\| V_1 \| \times \| V_2 \|}$\\
**** \\
}
As we are dealing with frequency vectors all coordinates are non negative. All vectors are therefore in the first orthant and the angles between them are in the range $[0,\pi/2]$ radians and therefore $cos\theta$ is in the range $[0,1]$ and so is $1-cos\theta$. \\
\textbf{Symmetry}: Both the dot product and the standard multiplication are commutative operations and therefore $ D(V_1,V_2)=\frac{V_1\cdot V_2}{\| V_1 \| \times \| V_2 \|}=\frac{V_2\cdot V_1}{\| V_2 \| \times \| V_1 \|}=D(V_2,V_1)$, and therefore also $1-D(V_1,V_2)=1-D(V_2,V_1)$.\\
\textbf{Identity of Indiscernibles}: 
First, we note that for vectors of length n, the standard dot product $V_1\cdot V_2$ is defined as $\sum_{i=1}^n V_{1_i}\times V_{2_i}$ and the standard euclidean norm is defined as $\sqrt{\sum_{i=1}^n (V_i)^2}$. 
Assume $V_1=V_2$, i.e. for every i $V_1(i)=V_2(i)$. We then have  $V_1\cdot V_2=\sum_{i=1}^n V_1(i)\times V_2(i) = \sum_{i=1}^n V_1(i)\times V_1(i) = \sum_{i=1}^n V_1(i)^2$.
We also have $\| V_1 \| = \|V_2\|$ and therefore $\|V_1\|\times\|V_2\|=\|V_1\|^2=\sqrt{\sum_{i=1}^n V_1(i)^2}^2=\sum_{i=1}^n V_1(i)^2$.
Therefore for $V_1=V_2$ we have $D(V_1,V_2)=\frac{V_1\cdot V_2}{\| V_1 \| \times \| V_2 \|}=\frac{\sum_{i=1}^n V_1(i)^2}{\sum_{i=1}^n V_1(i)^2}=1$ and $1-D(V_1,V_2)=0$.

Assume $1-D(V_1,V_2)=0$, that is $D(V_1,V_2)=1$. As we've seen when proving non-negativity, this implies that the angle $\theta$ between $V_1$ and $V_2$ is zero. Since both vectors are frequency vectors, i.e. $\|V_1\|=\|V_2\|=1$ this means they are the same vector.

\subsection*{L1 Norm}

\textbf{Non negativity:} It's enough to prove non negativity for each element of the sum, but that is assured by the absolute value. \\
\textbf{Symmetry:} Again, it's enough to show symmetry for each element of the sum. We note that $(V_1(x)-V_2(x))=-(V_2(x)-V_1(x))$ and therefore  $\arrowvert (V_1(x)-V_2(x))\arrowvert =\arrowvert (V_2(x)-V_1(x))\arrowvert $ \\
\textbf{Identity of indiscernibles:} Assume $V_1=V_2$, i.e for every $x\in X$ we have $V_1(x)=V_2(x)$. Then  $V_1(x)-V_2(x)=0$ and we have $L1(V_1,V_2)=0$.\\
Assume $V_1\neq V_2$ i.e. for some $x\in X$ we have $V_1(x)\neq V_2(x)$. For that $x$ we have $\big[V_1(x)-V_2(x)\big]>0$ and as all elements in the sum are non negative we have $L1(V_1,V_2)>0$.
\subsection*{KL divergence}
\textbf{Non negativity:} It is enough to show that each element in the sum is non negative. For every $x\in X$ either $V_1(x)<V_2(x)$, $V_1(x)V_2(x)$ or $V_1(x)=V_2(x)$. If $V_1(x)<V_2(x)$ then $V_1(x)-V_2(x)<0$ and $log\frac{V_1(x)}{V_2(x)}<0$. Therefore $\big[V_1(x)-V_2(x)\big]log\frac{V_1(x)}{V_2(x)}>0$. If $V_1(x)>V_2(x)$ then $V_1(x)-V_2(x)>0$ and $log\frac{V_1(x)}{V_2(x)}>0$ and again $\big[V_1(x)-V_2(x)\big]log\frac{V_1(x)}{V_2(x)}>0$. Lastly, if $V_1(x)=V_2(x)$ then $V_1(x)-V_2(x)=0$ and $log\frac{V_1(x)}{V_2(x)}=0$ and we have $\big[V_1(x)-V_2(x)\big]log\frac{V_1(x)}{V_2(x)}=0$ \\
\textbf{Symmetry:} We note that $(V_1(x)-V_2(x))=-(V_2(x)-V_1(x))$ and likewise $log\frac{V_1(x)}{V_2(x)}=-log\frac{V_2(x)}{V_1(x)}$ and therefore $(V_1(x)-V_2(x))log\frac{V_1(x)}{V_2(x)}=(V_2(x)-V_1(x))log\frac{V_2(x)}{V_1(x)}$.
Again, this is true for each element in the sum and therefore holds for the entire sum. \\
\textbf{Identity of indiscernibles:} Assume $V_1=V_2$, i.e for every $x\in X$ we have $V_1(x)=V_2(x)$. Then  $V_1(x)-V_2(x)=0$ and $log\frac{V_1(x)}{V_2(x)}=1$ and we have $\big[V_1(x)-V_2(x)\big]log\frac{V_1(x)}{V_2(x)}=0$. \\
Assume $V_1\neq V_2$ i.e. for some $x\in X$ we have $V_1(x)\neq V_2(x)$. For that $x$ we have $\big[V_1(x)-V_2(x)\big]log\frac{V_1(x)}{V_2(x)}>0$ and as all elements in the sum are non negative we have $D([V_1\Arrowvert V_2]=\sum_{x\in X}\Bigg[\big[V_1(x)-V_2(x)\big]log\Big[\frac{V_1(x)}{V_2(x)}\Big]\Bigg]>0$

\section{Datasets details}
\paragraph*{Books dataset: }	
The Books dataset is comprised of 30 English language books, taken from the Gutenberg project’s most popular books list. 
Each book is divided into chapters, varying from 7 to 61 chapters per book. We included only chapters that contain more than 150 words in the text and omitted words that appear less than five times in the book to avoid the extreme consequences of a very long tail.
Each text snippet is labeled with an author's name, and belongs to a book, and more specifically to a chapter in the book. In the Validation section, we use this subdivision to validate whether our method is able to distinguish between two texts written by the same person and two texts written by different people. We also test its ability to distinguish between a text written by one person and a text written by several, by collecting a number of chapters from different books to one virtual book. Table~\ref{tbl:books}in Appendix A lists the book names and relevant statistics.
\paragraph*{IMDb reviews dataset: }
\label{sec:imdbdata}
IMDb (Internet Movie Database) is among the world's most popular and authoritative sources for movie, TV and celebrity content. It offers a searchable database of more than 185 million data items, including more than 3.5 million films. This dataset contains 1,406,000 movie reviews, spanning the period of July 1998 - June 2016. 
We use their movie reviews text as our dataset for demonstrating the applications of our method. 
We define a user/author (contributor) as a registered person who published a movie review on IMDb. Each author is identified by a unique user id and alias. 
Each review contains a text, a timestamp, and an author ID. The original obtained IMDb dataset contained $467,961$ users. To have a large enough sample of text for each user, we extracted only users who published at least 30 reviews. 3,969 users met this criterion. We defined $n=30$ (number of different reviews per user) as our lower threshold to achieve statistical inference. The choice of $n = 30$ for a boundary between small and large samples is already used as a rule of thumb in many research areas: "The number 30 seems to have arisen from the understanding that with fewer than 30 cases, you were dealing with {\em small} samples that required specialized handling with {\em small-sample statistics} instead of the critical-ratio approach we have been taught"~\citep{cohen1990things}.
\begin{table*}[h!]
  \caption{Books datasets characteristics}
  \label{tbl:books}
  \begin{tabular}{ l l c c }
    \toprule
    \#&Book Name&	Unique &		Chapters	\\
    &~&Terms&(Avergae)\\
    \midrule
1&	A Study in Scarlet , Doyle, Arthur Conan&	16,388&	14\\
2&	A Tale of Two Cities , Dickens, Charles&		51,519&	39\\
3&	Adventures of Huckleberry Finn ,Twain, Mark&		39,066&	43\\
4&	Alice's Adventures in Wonderland , Carroll, Lewis&	9,066&	12	\\
5&	Anne of Green Gables,Montgomery, L. M. (Lucy Maud)&		41,005&	38\\
6&	Emma , Austen, Jane&		57,185&	55\\
7&	Great Expectations ,Dickens, Charles&		68,069&	59\\
8&	Grey Town ,Baldwin ,Gerald&		22,934&	24\\
9&	Gulliver Of Mars ,Arnold ,Edwin&		25,898&	27\\
10&	Gulliver's Travels, Swift, Jonathan&		39,117&	39\\
11&	Jane Eyre: An Autobiography ,Bront$\ddot{e}$, Charlotte&		76,612&	39\\
12&	Little Women ,Alcott, Louisa May&		73,266&	47\\
13&	Notes from the underground, Dostoevsky, Fyodor&		13,501&	14\\
14&	Oliver Twist ,Dickens, Charles&		61,895&	53\\
15&	Peter Pan ,Barrie, J. M. (James Matthew)&	17,095&	17\\
16&	Poll (The Lively), Ballantyne, R.M.&		12,240&	13\\
17&	Pride and Prejudice, Austen, Jane&		43,722&	61\\
18&	Sense and Sensibility, Austen, Jane&		42,478&	50\\
19&	The Adventures of Sherlock Holmes, Doyle, Arthur Conan&		34,559&	7\\
20&	The Adventures of Tom Sawyer , Twain, Mark&		27,077&	35\\
21&	The Devil-Tree of El Dorado, Aubrey, Frank&		39,409&	40\\
22&	The Following of the Star, Barclay ,Florence&		31,335&	32\\
23&	The King in Yellow, Chambers, Robert W. (Robert William)&		30,425&	28\\
24&	The Life and Adventures of Robinson Crusoe, Defoe, Daniel&		41,598&	20\\
25&	The Picture of Dorian Gray ,Wilde, Oscar&		29,947&	20\\
26&	The Time Machine ,Wells, H. G. (Herbert George)&	13,068&	12\\
27&	The Souls of Black Folk, DuBois, W.E.Burghardt&		27,046&	29\\
28&	Treasure Island ,Stevenson, Robert Louis&	25,451&	32\\
29&	Wuthering Heights ,Bront$\ddot{e}$, Emily&		47,183&	34\\
30&	Frey and His Wife, Hewlett, Maurice Henry&		9278&	10\\
	Total&~&		1,067,432&	943\\
  \bottomrule
\end{tabular}
\end{table*}
\begin{table*}[h!]
\centering
  \caption{IMDb dataset characteristics}
  \label{tbl:imdbrevs}
  \begin{tabular}{ l l}
    \toprule
    IMDb dataset property & Value\\ 
    \midrule
Start date&	July 1998\\
End date&	June 2016\\
Number of unique movies&	7,696\\
Users with more than 30 reviews&	3,969\\
Reviews&	474,961\\
Unique Nouns&	273,025\\
Avg unique nouns per user&	2,925\\
Avg number of reviews per movie&	61.7\\
Stdev number of reviews per movie&	71\\
Avg number of reviews per author&	119.6\\
Stdev number of reviews per author&	205\\
  \bottomrule
\end{tabular}
\end{table*}

\end{document}